\documentclass[11pt,a4paper]{article}
\usepackage[T1]{fontenc}
\usepackage{lmodern}
\usepackage[margin=25mm]{geometry}
\usepackage{amsmath,amssymb,amsfonts,graphicx,booktabs,array,multirow,xcolor}
\usepackage{placeins}
\usepackage[numbers,sort&compress]{natbib}
\usepackage[font=small,labelfont=bf,labelsep=space]{caption}
\usepackage[hidelinks]{hyperref}
\DeclareMathOperator*{\argmax}{arg\,max}
\DeclareMathOperator*{\argmin}{arg\,min}
\providecommand{\bibcommenthead}{}
\hypersetup{pdftitle={TME-ContextNet: Class-Conditioned Spatial Supervision and Grouped Decoding for Joint Nuclei Instance Segmentation and Classification},pdfauthor={Rukang Ma, Bingxue Wang, Liwen Zou, Yingying Cao, Zhenghua Cai, Yu Li, Dong Wang, Huiyu Zhou, Luying Gui},pdfsubject={Joint nuclei instance segmentation and classification},pdfkeywords={Computational pathology, nuclei instance segmentation, nuclei classification, spatial supervision, multi-task learning, efficient decoding}}
\title{\LARGE\bfseries TME-ContextNet: Class-Conditioned Spatial Supervision and Grouped Decoding for Joint Nuclei Instance Segmentation and Classification}
\newcommand{\preprintauthors}{\begin{minipage}{\dimexpr\textwidth-2\tabcolsep\relax}\centering
{\normalsize Rukang Ma\textsuperscript{1,7}\quad Bingxue Wang\textsuperscript{1}\quad Liwen Zou\textsuperscript{2}\quad Yingying Cao\textsuperscript{3}\quad Zhenghua Cai\textsuperscript{4}\\[4pt]
Yu Li\textsuperscript{1}\quad Dong Wang\textsuperscript{5,*}\quad Huiyu Zhou\textsuperscript{6,*}\quad Luying Gui\textsuperscript{1,*}\par}
\vspace{8pt}
{\footnotesize
\textsuperscript{1}School of Mathematics and Statistics, Nanjing University of Science and Technology, \mbox{Nanjing, 210094, China}\par
\textsuperscript{2}Department of Mathematics, Nanjing University, \mbox{Nanjing, 210093, China}\par
\textsuperscript{3}Department of Radiology, Affiliated Hospital of Nanjing University of Chinese Medicine, \mbox{Nanjing, 210018, China}\par
\textsuperscript{4}Department of Pancreatic and Bariatric Surgery, Nanjing Drum Tower Hospital, the Affiliated Hospital of Nanjing University Medical School, \mbox{Nanjing, 210008, China}\par
\textsuperscript{5}School of Artificial Intelligence and Information Technology, Nanjing University of Chinese Medicine, \mbox{Nanjing, 210023, China}\par
\textsuperscript{6}School of Computing and Mathematical Sciences, University of Leicester, \mbox{Leicester, LE1 7RH, United Kingdom}\par
\textsuperscript{7}School of Computer Science and Engineering, South China University of Technology, \mbox{Guangzhou, 510006, China}\par
}\vspace{5pt}
{\footnotesize\textsuperscript{*}Corresponding authors:\par
Dong Wang: \href{mailto:dongwang@seu.edu.cn}{dongwang@seu.edu.cn}\quad Huiyu Zhou: \href{mailto:hz143@leicester.ac.uk}{hz143@leicester.ac.uk}\par
Luying Gui: \href{mailto:ly.gui@njust.edu.cn}{ly.gui@njust.edu.cn}}
\end{minipage}}
\author{\preprintauthors}
\date{}
\begin{document}
\raggedbottom
\makeatletter
\setlength{\@fptop}{0pt}
\makeatother
\maketitle
\vspace{-1.6em}
\begin{abstract}
Joint nuclei instance segmentation and classification requires precise boundaries and neighborhood information that helps distinguish morphologically similar cell types. We present TME-ContextNet, a multi-task framework that learns class-conditioned spatial context from existing nuclear annotations and organizes decoding into geometry-related and semantic task groups. Class-conditioned spatial supervision extends nuclear labels into a dense target using Euclidean distance transforms, requiring no additional tissue-region annotation. We derive its nearest-region interpretation and characterize its stability to small annotation perturbations. A local--global context module combines dilated convolution and self-attention, while a grouped task decoder reduces redundant computation. The spatial head supplies training supervision and is omitted during inference. Experiments on PanMix, CoNSeP, and PanNuke yield Dice scores of 80.23\%, 65.32\%, and 79.13\%, and mean multi-class panoptic quality (mPQ+) scores of 58.67\%, 35.50\%, and 43.74\%, respectively. PanNuke results are averaged across three test folds. Target and decoder controls support the combination of distance-based supervision and grouped decoding while identifying their dependence on task configuration. The framework uses 36.40 million parameters, compared with 54.74 million for HoVer-Net. These findings support a compact, context-aware formulation for joint nuclei analysis in computational pathology.
\end{abstract}

\section{Introduction}\label{sec:intro}
Quantitative analysis of hematoxylin and eosin (H\&E) images requires identifying individual nuclei and assigning them biologically meaningful types. The resulting cellular objects support measurements of tissue composition, nuclear morphology, and spatial organization. These measurements are useful because tissue interpretation depends on both cellular appearance and the arrangement of different cell populations. For example, the spatial relationship between lymphocytes and tumor compartments provides information that a total nuclear count alone cannot express. Joint nuclei instance segmentation and classification is therefore a fundamental step in computational pathology and quantitative imaging \cite{sirinukunwattana2016locality,yuan2012quantitative,srinidhi2021deep}.

The joint task presents a tension between local precision and contextual recognition. Separating touching nuclei requires detailed boundaries and reliable instance geometry. Recognizing nuclear types can require a larger field of view because nuclei with similar appearance may occupy different cellular neighborhoods. Staining variation, heterogeneous morphology, and unequal class frequencies make these requirements harder to reconcile. A representation that improves semantic recognition must still retain the spatial detail needed to recover individual nuclei.

Current nuclei-analysis systems address these requirements through geometric prediction, contextual feature extraction, or interactions among multiple tasks. HoVer-Net combines foreground, displacement, and nuclear type prediction \cite{graham2019hover}; subsequent systems refine encoders, decoders, and whole-slide processing \cite{zhang2024keep,baumann2024hover}. These advances leave a specific opportunity in the supervision itself. Nuclear type annotations directly identify cell populations inside nuclear regions, while much of the surrounding non-nuclear space remains undifferentiated background. A larger receptive field exposes that space to the model but does not specify how existing cell annotations should organize its representation. At the same time, sharing all late decoder features can constrain tasks with different spatial requirements, whereas independent task decoders repeat substantial computation.

We investigate whether spatial supervision derived from nuclear classes can complement image context within a compact joint model. TME-ContextNet extends existing nuclear labels into the surrounding space through Class-conditioned Spatial Supervision (CSS). The implemented target uses Euclidean distance transforms (EDT) to associate non-nuclear locations with the nearest annotated class region, requiring no additional tissue-region annotation. A Local--Global Context Module (LGCM) combines dilated convolution with self-attention. A Grouped Task Decoder (GTD) shares intermediate features and then separates geometry-related outputs from semantic outputs. The CSS head shapes learning during training and is omitted from the inference outputs.

The contributions concern both the supervision and its integration. We formulate CSS as a class-conditioned spatial partition and derive the conditions under which its assignments are stable to small annotation perturbations. We introduce grouped decoding that couples CSS to nuclear type prediction while retaining a separate refinement path for nuclear geometry. We evaluate the complete framework on PanMix, CoNSeP, and PanNuke and examine target construction, component interactions, sharing depth, and computational cost. The complete model achieves Dice scores of 80.23\%, 65.32\%, and 79.13\%, respectively, and the highest mPQ+ among the evaluated specialist methods on each dataset. The controls characterize the contributions of spatial supervision, contextual features, and decoder sharing to the joint task.

\section{Related Work}\label{sec:related}
\subsection{Geometric representations and joint nuclear analysis}
Encoder--decoder architectures established the use of multi-resolution features for biomedical segmentation. U-Net combines contextual encoding with localization through skip connections \cite{ronneberger2015u}, and Micro-Net uses multiple image resolutions for microscopy segmentation \cite{raza2019micro}. Instance-oriented approaches add explicit object structure. StarDist predicts star-convex polygons \cite{schmidt2018stardist}, while Cellpose reconstructs instances from spatial flows \cite{stringer2021cellpose}. These representations address object separation, which is necessary before reliable cell-level quantities can be measured.

Joint nuclear analysis adds the requirement that each separated object receive a type. HoVer-Net predicts foreground, horizontal--vertical displacement, and nuclear type in a common framework \cite{graham2019hover}. SONNET uses self-guided ordinal regression to stratify nuclear regions \cite{sonnet2022}, and PointNu-Net connects semantic keypoints to dynamically generated instance masks \cite{pointnu2024}. SMILE studies cost-sensitive multi-task learning and task interaction under class imbalance \cite{pan2023smile}. Together, these approaches establish that nuclear geometry and semantic identity can benefit from coordinated learning, while leaving several choices for the outputs and their supervision.

\subsection{Context and annotation-derived spatial supervision}
Contextual methods exploit neighboring cells and tissue appearance to interpret nuclear morphology \cite{sirinukunwattana2016locality,nguyen2015using,bai2022context}. Context and attention also recur across the recent literature on nuclei instance segmentation and classification \cite{nunes2025survey}. Transformer-based models provide another mechanism for relating local structures to a wider field of view. CellViT combines a vision-transformer encoder with task-specific decoding for joint segmentation and classification \cite{hoer2024cellvit}, while CellViT++ adapts foundation-model representations to changing cell taxonomies through separate segmentation and classification stages \cite{hoer2026cellvitpp}. CFR-SAM introduces local feature adaptation and fine-grained boundary refinement for nuclei analysis \cite{cfrsam2026}.

Auxiliary targets provide a complementary way to organize learning. CAS-UNet uses boundary-distance supervision for nuclear-region segmentation \cite{casunet2026}. Its distance target describes nuclear geometry, whereas CSS uses class-specific nuclear supports to assign the surrounding non-nuclear space to cell populations. The distinction concerns what the auxiliary output represents: an instance- or boundary-related geometric quantity and a class-conditioned neighborhood partition impose different constraints on a shared representation. CSS requires no additional region annotations: it uses existing nuclear labels to construct class-conditioned spatial supervision. Its geometric target makes cellular neighborhood organization available to representation learning.

\subsection{Decoder sharing and computational efficiency}
Recent specialist methods also improve the architecture and processing cost of joint analysis. Zhang et al. combine enhanced encoders, nested dense decoding, regularization, and ensemble strategies \cite{zhang2024keep}. HoVer-NeXt emphasizes efficient nuclei analysis within a whole-slide pipeline \cite{baumann2024hover}, and NuLite develops lightweight joint models \cite{nulite2026}. These studies make computational cost and the organization of decoding important comparison dimensions. GTD studies a particular division of the late representation: foreground and displacement form a geometry group, while nuclear type and CSS form a semantic group. The corresponding controls compare this grouping with independent and plain shared decoding.

General-purpose segmentation architectures provide a different reference for nuclear class-map prediction. PSPNet aggregates context through pyramid pooling \cite{zhao2017pyramid}; K-Net uses iteratively updated kernels for segmentation \cite{zhang2021knet}; Mask2Former applies masked attention to segmentation \cite{cheng2022masked}; and SegFormer combines a hierarchical transformer encoder with lightweight decoding \cite{xie2021segformer}. They are included as representative segmentation architectures with different ways of using context. Their class maps test whether the proposed framework improves pixel-level nuclear type assignments beyond established general segmentation designs. Specialist nuclei-analysis methods provide the complementary comparison for correctly delineated and typed instances. Keeping these roles explicit makes the pixel and instance results answer distinct questions.

\section{Materials and Methods}\label{sec:methods}
\subsection{Task and framework}\label{sec:formulation}
Let $X\in\mathbb{R}^{H\times W\times3}$ be an H\&E image on a pixel domain $\Omega$. The annotations consist of an instance map $I$ and a nuclear type map $Y\in\{0,1,\ldots,C\}^{H\times W}$, where zero denotes background. The required outputs are an instance map $\widehat I$ and one type label for each predicted nucleus. During training, the network additionally predicts nuclear foreground probabilities $\widehat N$, horizontal--vertical displacement maps $\widehat V$, nuclear type probabilities $\widehat P$, and an auxiliary spatial target $\widehat T$.

\begin{figure}[tbp]
\centering\includegraphics[width=\linewidth]{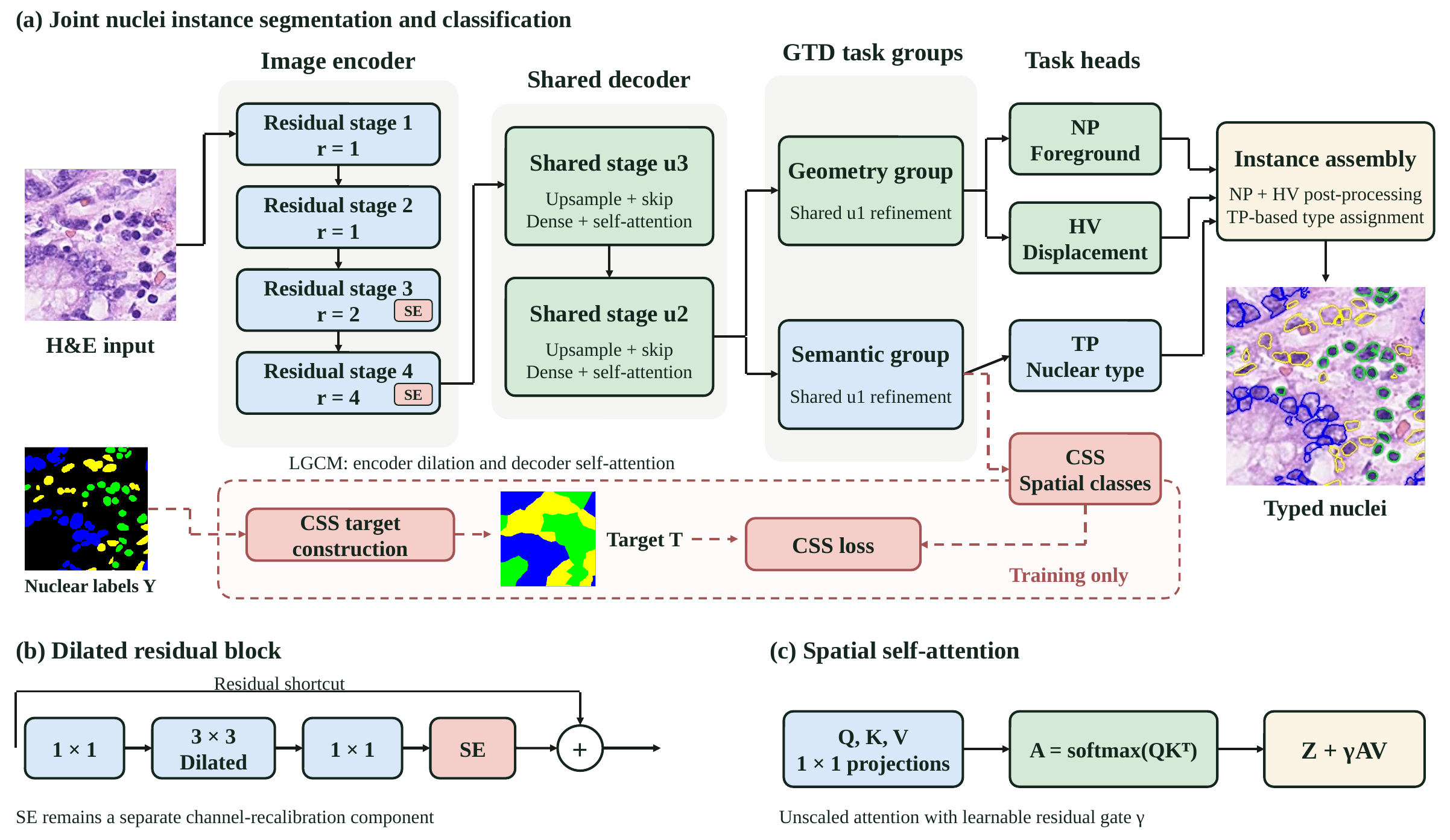}
\caption{Overview of TME-ContextNet. (a) The image encoder and shared decoder feed geometry-related and semantic task groups. Nuclear foreground (NP), horizontal--vertical displacement (HV), and nuclear type (TP) predictions are used for typed instance assembly. Class-conditioned spatial supervision (CSS) provides an annotation-derived auxiliary training target, and its head is omitted at inference. Encoder dilation and decoder self-attention form the local--global context module (LGCM), while squeeze-and-excitation (SE) supplies separate channel recalibration. (b) Dilated residual block with a shortcut connection. (c) Spatial self-attention with a learnable residual gate. GTD denotes the grouped task decoder}
\label{fig:framework}
\end{figure}

An encoder produces multi-resolution features $E(X)$. Shared upsampling stages combine these features before the decoder separates them into two groups: a geometry group for foreground and displacement, and a semantic group for nuclear type and CSS. Figure~\ref{fig:framework} summarizes the computation. The CSS target is generated from the training annotations. At test time, the network takes the image alone; foreground and displacement predictions produce instances through HoVer-style post-processing, and type predictions assign a label to each instance \cite{graham2019hover}.

\subsection{Class-conditioned spatial supervision (CSS)}\label{sec:css}
\begin{figure}[tbp]
\centering\includegraphics[width=\linewidth]{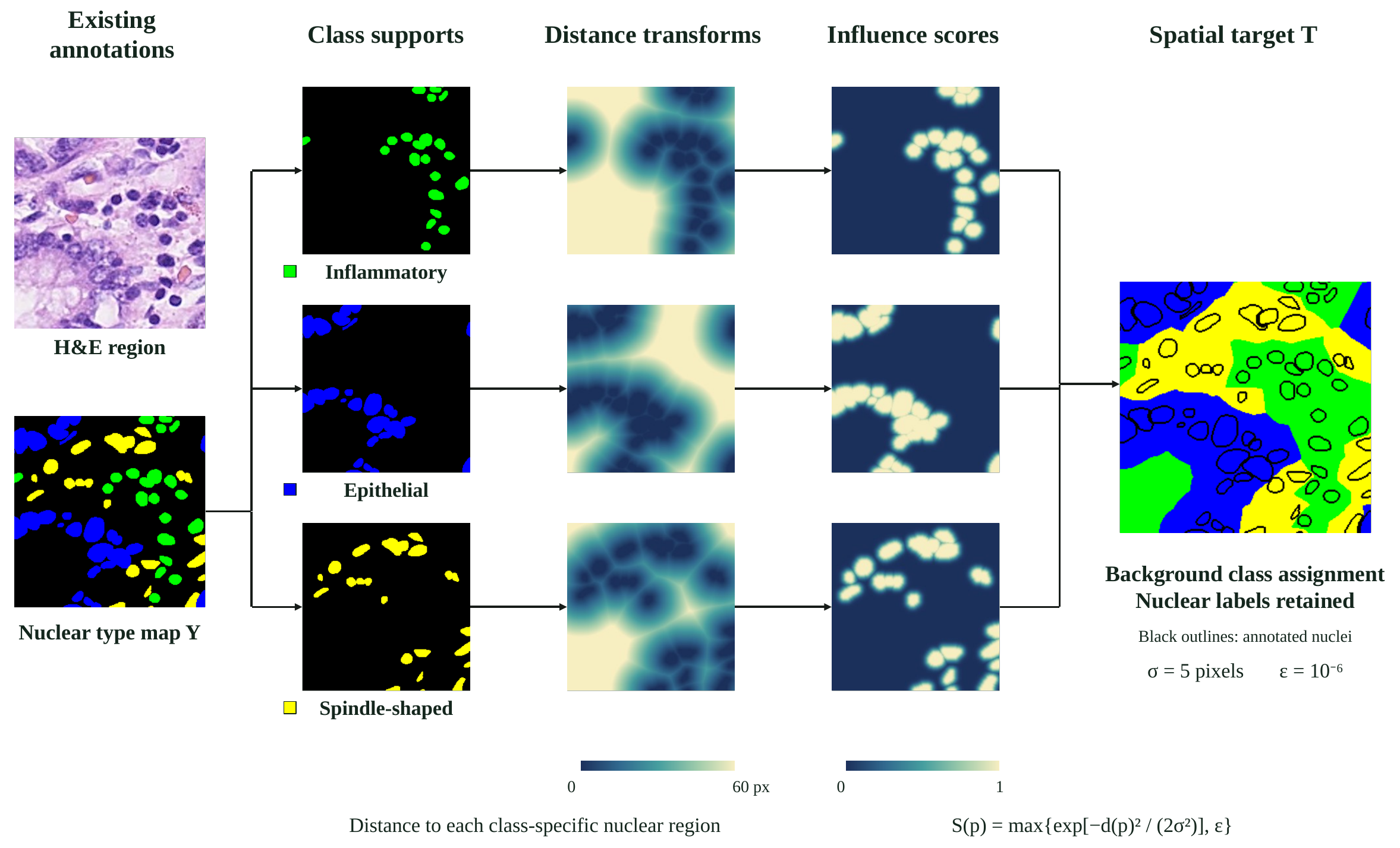}
\caption{Construction of the EDT-based CSS target. The nuclear type map is separated into class-specific supports. Each support produces a Euclidean distance transform and an influence score with $\sigma=5$ pixels and $\varepsilon=10^{-6}$. Background pixels receive the class with the largest score, while annotated nuclei retain their original labels. Black outlines identify nuclear regions on the final target. Distance and score maps use continuous color scales, and categorical maps use the corresponding nuclear class colors}
\label{fig:cssgen}
\end{figure}

For every class $c$ present in an annotated image, define its nuclear support and distance transform by
\begin{equation}
 F_c=\{p\in\Omega:Y(p)=c\},\qquad
 d_c(p)=\min_{q\in F_c}\|p-q\|_2 .\label{eq:distance}
\end{equation}
Thus, distance is measured to the class-specific nuclear region rather than to a nuclear centroid. Let $\mathcal C_X=\{c:F_c\ne\varnothing\}$. The implemented influence score is
\begin{equation}
 S_c(p)=\max\left\{\exp\!\left[-\frac{d_c(p)^2}{2\sigma^2}\right],\varepsilon\right\},
 \quad c\in\mathcal C_X,\label{eq:score}
\end{equation}
with a common scale $\sigma=5$ pixels and numerical floor $\varepsilon=10^{-6}$. The target preserves the annotated nuclear labels and extends the class assignments into background:
\begin{equation}
 T(p)=\begin{cases}
 Y(p),&Y(p)>0,\\
 \underset{c\in\mathcal C_X}{\arg\max}\;S_c(p),&Y(p)=0,\ \mathcal C_X\ne\varnothing,\\
 0,&\mathcal C_X=\varnothing.
 \end{cases}\label{eq:target}
\end{equation}
Ties are resolved by the first class in the ordered list of present classes, as in the generation code. Figure~\ref{fig:cssgen} illustrates the construction.

\paragraph{Geometric interpretation.}
When the largest influence is unique and above the numerical floor, the common exponential in Eq.~\eqref{eq:score} is strictly decreasing in distance. Consequently,
\begin{equation}
 \argmax_c S_c(p)=\argmin_c d_c(p).
 \label{eq:nearest}
\end{equation}
The EDT target is therefore a class-conditioned nearest-region partition of the non-nuclear space. It gives adjacent cells a shared contextual region when they belong to the same class while maintaining a transition between neighborhoods of different classes. Nuclear appearance supplies the image evidence, and the spatial partition supplies an additional organization of the supervision.

The common $\sigma$ changes the continuous scores but, away from numerical-floor ties, does not change the hard class assignments. The main construction consequently does not require a set of class-specific influence strengths. The auxiliary signal expresses spatial proximity to annotated cell populations. Its biological motivation is the relevance of cellular organization to tissue interpretation \cite{yuan2012quantitative}; its mathematical content is a geometric representation derived from those annotations.

The construction also has a useful stability property. If a class support changes from $F_c$ to $F'_c$ by at most $\delta$ in Hausdorff distance, then $|d_c(p)-d'_c(p)|\leq\delta$. A winning nearest class therefore remains unchanged wherever its distance margin over the runner-up exceeds $2\delta$, provided that numerical clipping does not produce a tie. Appendix~\ref{app:derivation} gives the derivation. This property characterizes the stability of the spatial target under small annotation perturbations.

\subsection{Local--global context module (LGCM)}\label{sec:lgcm}
LGCM combines dilated residual feature extraction in the encoder with self-attention inside decoder dense blocks. Residual learning and atrous convolution provide the basis for its convolutional context extraction \cite{he2016deep,chen2018encoder}, while dense connections support feature reuse within the decoder \cite{huang2017densely}. Together, these operations provide local and non-local image context. SE remains a separate channel-recalibration component. For a residual unit with input $Z_{k-1}$, write
\begin{equation}
 Z_k=\mathcal A_k\!\left(\mathcal G_k(Z_{k-1};r_k)\right)+\mathcal P_k(Z_{k-1}),
 \label{eq:residual}
\end{equation}
where $\mathcal G_k$ is a bottleneck convolutional transformation with dilation $r_k$, $\mathcal P_k$ aligns shortcut channels and resolution, and $\mathcal A_k$ denotes the spatial alignment used before addition. Dilation increases the spatial sampling extent while retaining the convolutional kernel's parameter count. With kernel width $k_l$, dilation $r_l$, stride $s_l$, receptive-field width $R_l$, and sampling jump $j_l$, the standard recurrence is
\begin{equation}
 R_l=R_{l-1}+(k_l-1)r_lj_{l-1},\qquad
 j_l=s_lj_{l-1}.\label{eq:rf}
\end{equation}
This relationship describes theoretical convolutional support. Self-attention then enables feature exchange across the spatial locations represented at the attention stage.

For a flattened feature matrix $Z\in\mathbb R^{n\times d}$, let $Q=ZW_Q$, $K=ZW_K$, and $V=ZW_V$. Spatial aggregation follows the non-local attention formulation \cite{wang2018nonlocal}. The implemented attention and residual output are
\begin{equation}
 A=\operatorname{softmax}_{\rm row}(QK^\top),\qquad
 Z'=Z+\gamma AV.\label{eq:attention}
\end{equation}
The learnable gate $\gamma$ is initialized to zero, so the block initially follows its local feature path and learns the contribution of non-local aggregation during optimization. Equation~\eqref{eq:attention} follows the unscaled dot-product attention in the implementation. Squeeze-and-excitation (SE) applies channel recalibration \cite{hu2018squeezeandexcitation} and is controlled separately from LGCM in the experiments. This separation allows the contribution of dilation and spatial attention to be examined without conflating it with channel gating.

\subsection{Grouped task decoder (GTD)}\label{sec:gtd}
GTD uses a shared decoder representation followed by group-specific transformations and separate prediction heads. Let $Z=D_{\rm sh}(E(X))$ be the shared representation. Its outputs are
\begin{align}
 (\widehat N,\widehat V)&=(h_N(G_{\rm geom}(Z)),h_V(G_{\rm geom}(Z))),\\
 (\widehat P,\widehat T)&=(h_P(G_{\rm sem}(Z)),h_T(G_{\rm sem}(Z))).
 \label{eq:groups}
\end{align}
The grouping reflects the outputs' roles: foreground and displacement concern nuclear geometry, whereas type and CSS assign semantic classes at different spatial supports. The decoder shares intermediate features rather than treating every task as either fully independent or identical throughout decoding.

The parameter saving can be expressed without assuming a particular channel width. Suppose $m$ independent task decoders each contain a trunk with $P_D$ parameters and a head with $P_{h_t}$ parameters. For an encoder with $P_E$ parameters,
\begin{align}
 P_{\rm independent}&=P_E+mP_D+\sum_tP_{h_t},\\
 P_{\rm grouped}&=P_E+P_{\rm sh}+\sum_gP_{G_g}+\sum_tP_{h_t},\\
 \Delta P&=mP_D-P_{\rm sh}-\sum_gP_{G_g}.
 \label{eq:parameters}
\end{align}
Sharing is parameter-efficient when $\Delta P>0$. The accuracy associated with a particular sharing arrangement depends on the tasks and supervision, which motivates the decoder controls in Table~\ref{tab:structures}. Appendix Fig.~\ref{fig:merging} illustrates the decoder structures and sharing stages.

During inference, $h_T$ is omitted. In Eq.~\eqref{eq:groups}, the retained outputs do not use its predicted value, so omitting the auxiliary head preserves their forward computation for fixed shared weights. CSS can thus alter the representation learned during training without adding a required prediction stage to the deployed nuclei-analysis output.

\subsection{Training objective and auxiliary gradient}\label{sec:loss}
Foreground and type prediction use cross-entropy and soft Dice losses following HoVer-Net \cite{graham2019hover}. For one-hot targets $y_{uc}$ and class probabilities $\pi_{uc}$ over supervised pixels $u\in\Omega_s$, define
\begin{equation}
\begin{aligned}
 \mathcal L_{\rm CE}(y,\pi)&=-\frac1{|\Omega_s|}\sum_{u\in\Omega_s}\sum_cy_{uc}\log\pi_{uc},\\
 \mathcal L_{\rm Dice}(y,\pi)&=\sum_c\left[1-\frac{2\sum_uy_{uc}\pi_{uc}+\epsilon_D}
 {\sum_uy_{uc}+\sum_u\pi_{uc}+\epsilon_D}\right].
\end{aligned}\label{eq:losses}
\end{equation}
The classification branches use softmax cross-entropy with Dice, consistent with mutually exclusive type targets. Foreground prediction uses the corresponding binary two-class form. Displacement supervision combines mean squared error with a foreground-focused gradient error. The total objective is
\begin{equation}
 \mathcal L=\mathcal L_N+\mathcal L_V+\mathcal L_P+\omega\mathcal L_T,\qquad
 \mathcal L_T=\mathcal L_{\rm CE}(T,\widehat T)+\mathcal L_{\rm Dice}(T,\widehat T),
 \label{eq:total}
\end{equation}
where branch coefficients may equivalently be absorbed into the individual loss terms and $\omega$ controls auxiliary supervision. For the controlled CoNSeP target experiments, both auxiliary loss terms have coefficient $0.5$ and both primary classification loss terms have coefficient one; the no-CSS condition sets the auxiliary coefficient to zero. The CSS head supplies gradients to the shared representation even though its output is omitted at inference.

The choice and weighting of auxiliary objectives matter in multi-task learning: uncertainty-based weighting addresses relative loss scales \cite{kendall2018uncertainty}, while gradient surgery addresses interference between task updates \cite{yu2020pcgrad}. Here, the focus is the spatial target supplied to the shared representation. To make this role explicit, let $g=\nabla_\theta\mathcal L_{\rm main}$ and $a=\nabla_\theta\mathcal L_T$ for shared parameters $\theta$, where $\mathcal L_{\rm main}=\mathcal L_N+\mathcal L_V+\mathcal L_P$. For fixed $\omega$, assume that the primary loss has a locally Lipschitz gradient near the current shared parameters. Under a small gradient step $\theta'=\theta-\eta(g+\omega a)$, a first-order expansion gives
\begin{equation}
 \mathcal L_{\rm main}(\theta')-\mathcal L_{\rm main}(\theta)
 =-\eta\|g\|^2-\eta\omega\langle g,a\rangle+O(\eta^2).
 \label{eq:gradient}
\end{equation}
The auxiliary contribution $-\eta\omega\langle g,a\rangle$ favors the immediate primary-loss decrease when $\omega>0$ and the gradients align. A negative inner product reduces that decrease, although the primary term can still dominate. The remainder is of order $\eta^2$ for a fixed update direction; Appendix~\ref{app:derivation} gives a finite-step bound. Equation~\eqref{eq:gradient} motivates testing alternative spatial targets rather than assuming that any dense target is beneficial. The target controls evaluate this design choice through observed performance; gradient alignment itself was not measured in those experiments.

\subsection{Datasets and experimental protocol}\label{sec:protocol}
The study uses PanMix, CoNSeP, and PanNuke. PanMix is a private pancreatic histopathology collection with annotations for tumor, lymphocyte, fibroblast, acinar, and islet nuclear categories. Access is organized through an application procedure described on the associated GitHub page. CoNSeP provides colorectal H\&E images with nuclear instance annotations; the analysis uses the four-category grouping associated with joint nuclei analysis \cite{graham2019hover}. PanNuke contains diverse tissue types and five nuclear classes: neoplastic, inflammatory, connective, dead, and epithelial \cite{gamper2019pannuke,gamper2020pannuke}.

\begin{figure}[tbp]
\centering\includegraphics[width=\linewidth]{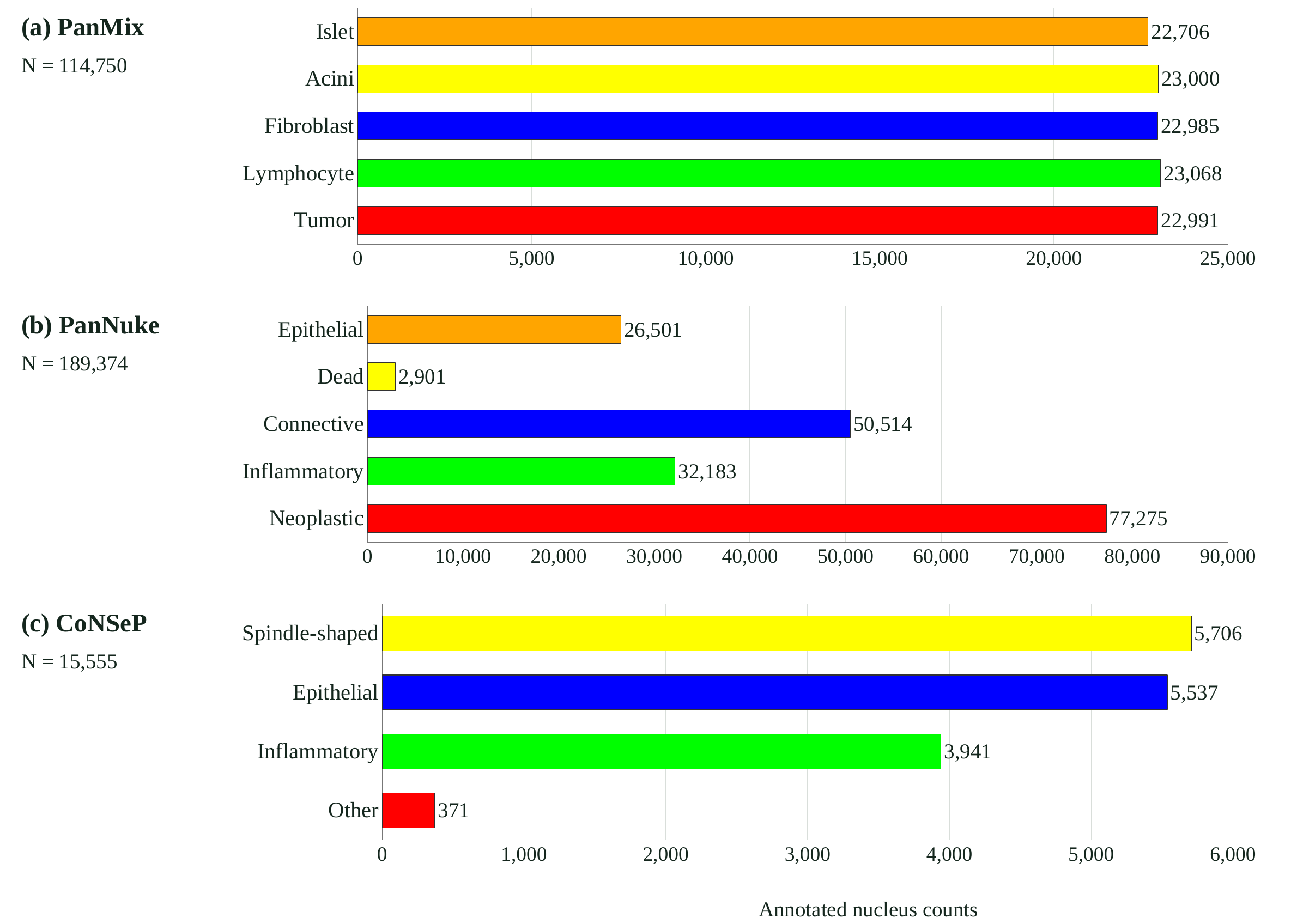}
\caption{Nuclear class composition from the PanMix and PanNuke dataset statistics and the CoNSeP training partition. Bars show annotated nucleus counts from the dataset statistics. Colors follow the class order used for each dataset in the qualitative figures. Axis scales differ among panels}
\label{fig:composition}
\end{figure}

Figure~\ref{fig:composition} summarizes the annotated nuclear class counts from the dataset statistics; the CoNSeP panel corresponds to its training partition. PanMix has similar counts across its five categories, whereas PanNuke contains a much smaller dead-cell population than its neoplastic and connective populations. The grouped CoNSeP labels also have unequal counts. This composition provides context for interpreting overall overlap and the class-specific results. The dataset-specific taxonomies are retained throughout evaluation.

All models were trained without pretrained weights for 50 epochs, and the best checkpoint was selected using the validation set. PanNuke was evaluated with three fold rotations, each providing an independent test fold. PanMix and CoNSeP use the established training, validation, and test partitions of the experiments. Test data were reserved for final evaluation. For TME-ContextNet, the implementation uses Adam with an initial learning rate of $10^{-4}$. The controlled CoNSeP experiments use batch size four, Adam coefficients $(0.9,0.999)$, and a tenfold learning-rate reduction every 25 epochs. Their validation criterion is the sum of the foreground, displacement, and type losses, excluding the training-only auxiliary loss. Input crops are $270\times270$ pixels with an $80\times80$ supervised central region. Image preprocessing and augmentation follow the established experimental pipeline.

The comparison has two complementary scopes. PSPNet, K-Net, Mask2Former, and SegFormer \cite{zhao2017pyramid,zhang2021knet,cheng2022masked,xie2021segformer} evaluate the pixel classification component using their predicted nuclear class maps. Their selection spans pyramid context, kernel-based segmentation, masked attention, and hierarchical transformer encoding. HoVer-Net, Zhang et al., HoVer-NeXt, and CellViT \cite{graham2019hover,zhang2024keep,baumann2024hover,hoer2024cellvit} additionally evaluate joint typed-instance prediction. CellViT adds a vision-transformer specialist to the convolutional and ConvNeXt-based comparators. The Zhang implementation uses a SEResNeXt-50 encoder, and the HoVer-NeXt implementation uses ConvNeXtV2-Tiny. All reported comparisons use the trained implementations under the 50-epoch, random-initialization protocol described above. CellViT uses a trainable ViT-S encoder with embedding dimension 384, 12 transformer blocks, six attention heads, and feature extraction at blocks 3, 6, 9, and 12. It retains the nuclear foreground, displacement, and type branches and their original loss coefficients. Training uses $256\times256$ patches, batch size four, FP32, and AdamW with an initial learning rate of $3\times10^{-4}$, weight decay $10^{-4}$, and cosine decay. Its tissue-classification auxiliary loss is disabled because the prepared training manifests supply nuclear labels only; patches are sampled uniformly. The validation nuclear-branch loss selects the checkpoint, and native HoVer-style post-processing produces typed instances.

All methods contribute Dice and IoU where a nuclear class map is available. mPQ+ additionally requires nuclear instance identities and is reported for the evaluated specialist outputs. Dashes mark outputs without nuclear instance identities. Appendix Fig.~\ref{fig:semanticcomparison} supplies qualitative class-map comparisons for the general segmentation methods, while Fig.~\ref{fig:comparison} focuses on specialist joint analysis.

The controls serve separate questions. PanMix component experiments examine the configurations in Table~\ref{tab:components}. To isolate the construction of CSS, the CoNSeP target experiments use a dedicated HoVer-Net-based control model with an auxiliary spatial prediction head. They compare EDT with no auxiliary target, Gaussian constructions, class-weight controls, and a class-agnostic distance target. This target-design experiment is reported separately from the complete TME-ContextNet comparison in Table~\ref{tab:overall}. The no-CSS and EDT conditions each use seeds 10, 20, and 30; the remaining target conditions use seed 10. The auxiliary head architecture is retained across these target controls, and its loss is disabled in the no-CSS condition. Appendix~\ref{app:controls} specifies each target. The CoNSeP decoder and LGCM comparisons use seed 10 and are organized into the matched comparison blocks in Table~\ref{tab:structures}. The PanMix sharing-depth experiment is a controlled comparison: CSS, SE, and self-attention are disabled in all three stages, and all remaining training and evaluation settings are held fixed. The only variable is the extent of decoder sharing: Stage I shares u3, Stage II shares u3 and u2, and Stage III shares u3, u2, and u1.

The archived auxiliary-weight sensitivity experiment uses PanMix and the EDT target, varying $\omega$ from 0.0 to 1.0 in increments of 0.1. Each weight uses one training run with the remaining settings held fixed; each boxplot summarizes scores across test images. This experiment addresses the strength of spatial supervision, whereas the CoNSeP controls address its construction. The Dice and IoU boxplots describe the per-image response to auxiliary-loss strength.

For efficiency measurements, network latency denotes the timed forward computation, whereas image latency includes image assembly and post-processing and excludes disk input/output. The benchmark uses 32-bit floating-point precision (FP32) with batch size one. Image and P95 latency refer to a $256\times256$ image; GFLOPs refer to one native network input patch. All latency and memory measurements in Table~\ref{tab:efficiency} use RTX 4090. The main-model and decoder blocks identify separate measurement sessions, with models within each block evaluated sequentially on the same GPU.

\subsection{Evaluation and reporting}\label{sec:metrics}
Dice and intersection over union (IoU) follow the pixel-level evaluator used for the result collection. For image $i$, let $G_{ic}$ and $P_{ic}$ be the ground-truth and predicted regions of class $c$. The evaluator computes
\begin{align}
 \operatorname{Dice}_{ic}&=\frac{2|G_{ic}\cap P_{ic}|+\epsilon_m}{|G_{ic}|+|P_{ic}|+\epsilon_m},\\
 \operatorname{IoU}_{ic}&=\frac{|G_{ic}\cap P_{ic}|+\epsilon_m}{|G_{ic}\cup P_{ic}|+\epsilon_m},
 \qquad \epsilon_m=10^{-5}.\label{eq:overlap}
\end{align}
The overall pixel score is the unweighted mean across images and classes $c=0,\ldots,C$, including background. With this smoothing, an empty ground-truth and predicted class mask scores one. The reported Dice and IoU therefore measure class-map overlap.

Mean multi-class panoptic quality (mPQ+) evaluates type-aware instances, building on the panoptic quality formulation \cite{kirillov2019panoptic} and its use in joint nuclei analysis \cite{graham2019hover}. For foreground instances of class $c$, matches require the same class and IoU strictly greater than $0.5$. With matched pairs $TP_c$ and unmatched predictions and annotations $FP_c$ and $FN_c$, respectively,
\begin{align}
 \operatorname{DQ}_c&=\frac{|TP_c|}{|TP_c|+\tfrac12|FP_c|+\tfrac12|FN_c|},\\
 \operatorname{SQ}_c&=\frac{\sum_{(g,p)\in TP_c}\operatorname{IoU}(g,p)}{|TP_c|},\\
 \operatorname{PQ}^{+}_c&=\frac{\sum_{(g,p)\in TP_c}\operatorname{IoU}(g,p)}
 {|TP_c|+\tfrac12|FP_c|+\tfrac12|FN_c|}
 =\operatorname{DQ}_c\operatorname{SQ}_c,\\
 \operatorname{mPQ}^{+}&=\frac1{|\mathcal C_{\rm eval}|}\sum_{c\in\mathcal C_{\rm eval}}\operatorname{PQ}^{+}_c.
 \label{eq:pq}
\end{align}
Here, DQ denotes detection quality and SQ denotes segmentation quality. The class-wise matching statistics are accumulated across the evaluation set before calculating each $\operatorname{PQ}^{+}_c$. The set $\mathcal C_{\rm eval}$ contains foreground classes with a nonzero PQ denominator; undefined empty/empty classes are omitted. SQ is set to zero when there are no matched instances. This metric evaluates type-aware detection and instance delineation together. Table~\ref{tab:classes} additionally reports the product of smoothed pixel precision and recall, defined for each image by
\begin{equation}
 \operatorname{PR}_{ic}=\frac{(|G_{ic}\cap P_{ic}|+\epsilon_m)^2}
 {(|P_{ic}|+\epsilon_m)(|G_{ic}|+\epsilon_m)}.
\end{equation}
The product is averaged across images and labeled Pixel $P\times R$ in the class-wise table. Instance-level quality is measured by mPQ+ in Eq.~\eqref{eq:pq}.

PanNuke values are reported as the mean and sample standard deviation across its three test folds. For the repeated CSS conditions, Table~\ref{tab:targets} reports mean and sample standard deviation across three random seeds; the remaining target conditions are single-run results. For measurements $z_1,\ldots,z_n$, the mean and sample standard deviation are $\bar z=n^{-1}\sum_i z_i$ and $s=[\sum_i(z_i-\bar z)^2/(n-1)]^{1/2}$, respectively. Single-run results are shown without an estimated standard deviation. Comparisons are descriptive.

\section{Results}\label{sec:results}
\subsection{Joint analysis across three datasets}\label{sec:overall}
Table~\ref{tab:overall} reports the complete comparison. TME-ContextNet achieves the highest Dice, IoU, and mPQ+ values on all three datasets. On PanMix, its Dice and mPQ+ are 80.23\% and 58.67\%, respectively. On CoNSeP, the corresponding values are 65.32\% and 35.50\%. On PanNuke, it obtains Dice of $79.13\pm0.80$\%, IoU of $74.85\pm0.81$\%, and mPQ+ of $43.74\pm2.89$\% across the three test folds.

The pixel-level comparison is strongest when measured against the best general segmentation result for each dataset. TME-ContextNet exceeds K-Net by 1.06 Dice points on PanMix and 0.67 points on PanNuke, and exceeds Mask2Former by 1.16 points on CoNSeP. These consistent overlap differences extend the comparison to established general segmentation architectures. The general segmentation models remain competitive, particularly on the overlap measures for which they produce directly comparable class maps.

The specialist baselines provide the instance-level comparison. HoVer-Net has the strongest competing mPQ+ on PanMix, Zhang et al. on CoNSeP, and CellViT on PanNuke. The corresponding margins for TME-ContextNet are 10.82, 1.10, and 1.83 percentage points. CellViT achieves $76.33\pm0.38$\% Dice, $72.04\pm0.36$\% IoU, and $41.91\pm0.30$\% mPQ+ on PanNuke. TME-ContextNet exceeds these means by 2.80, 2.81, and 1.83 points, respectively. CellViT also obtains Dice/mPQ+ of 73.51\%/41.09\% on PanMix and 57.76\%/29.74\% on CoNSeP, providing a transformer-based reference for the complete joint task.

The smaller CoNSeP margin and larger PanMix margin show that the size of the advantage depends on the dataset and metric. Pixel overlap and mPQ+ should therefore be read together: a plausible class region can still contain a merged nucleus, a missed object, or an incorrect instance type.

\begin{figure}[tbp]
\centering\includegraphics[width=\linewidth]{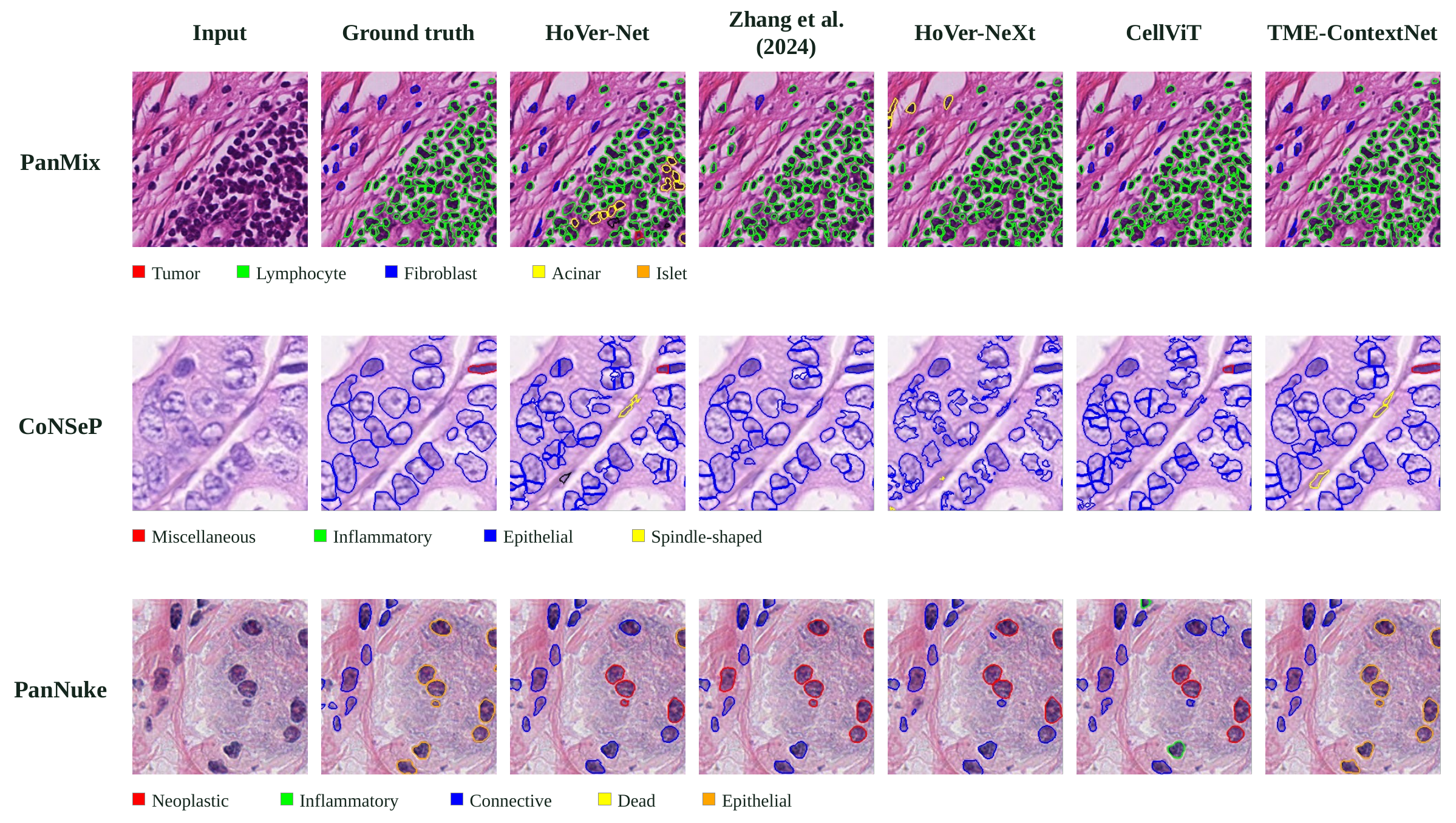}
\caption{Qualitative examples of joint nuclei instance segmentation and classification on PanMix, CoNSeP, and PanNuke. Columns show the H\&E input, ground truth, HoVer-Net, Zhang et al. (2024), HoVer-NeXt, CellViT, and TME-ContextNet. Identical image regions and class colors are used across methods within each dataset. Colored contours indicate individual nuclei and their assigned types, with dataset-specific class legends below each row}
\label{fig:comparison}
\end{figure}

Figure~\ref{fig:comparison} compares all five specialist methods at identical image coordinates. The PanMix field combines a dense lymphocyte region with elongated fibroblast nuclei, exposing differences in both object separation and class assignment. In the CoNSeP field, the larger epithelial nuclei reveal fragmentation in several predictions, including CellViT. In the PanNuke field, CellViT and the other baselines assign several annotated epithelial nuclei to the neoplastic class, while TME-ContextNet preserves their epithelial assignments. These examples connect the joint instance metric to visible differences in contours and types. The supplementary class-map comparison in Appendix Fig.~\ref{fig:semanticcomparison} addresses the general segmentation baselines separately.
\begin{table}[!htbp]
\centering
\small
\caption{Comparison on PanMix, CoNSeP, and PanNuke. Values are percentages; bold indicates the highest value in each dataset and metric.}
\label{tab:overall}
\setlength{\tabcolsep}{4pt}
\renewcommand{\arraystretch}{1.12}
\begin{tabular}{lccc}
\toprule
\multicolumn{4}{l}{\textbf{PanMix}} \\
Method & Dice (\%) & IoU (\%) & mPQ+ (\%) \\
\midrule
PSPNet & 75.62 & 73.04 & --- \\
KNet & 79.17 & 76.01 & --- \\
Mask2Former & 77.77 & 75.67 & --- \\
Segformer & 76.18 & 73.49 & --- \\
HoVer-Net & 69.80 & 67.22 & 47.85 \\
Zhang et al. (2024) & 71.27 & 68.85 & 42.56 \\
HoVer-NeXt & 69.85 & 67.18 & 29.16 \\
CellViT & 73.51 & 70.85 & 41.09 \\
TME-ContextNet & \textbf{80.23} & \textbf{78.52} & \textbf{58.67} \\
\midrule
\multicolumn{4}{l}{\textbf{CoNSeP}} \\
Method & Dice (\%) & IoU (\%) & mPQ+ (\%) \\
\midrule
PSPNet & 54.66 & 46.04 & --- \\
KNet & 55.91 & 48.13 & --- \\
Mask2Former & 64.16 & 55.33 & --- \\
Segformer & 60.79 & 51.67 & --- \\
HoVer-Net & 49.94 & 41.49 & 23.35 \\
Zhang et al. (2024) & 59.02 & 51.24 & 34.40 \\
HoVer-NeXt & 45.94 & 37.51 & 20.51 \\
CellViT & 57.76 & 49.07 & 29.74 \\
TME-ContextNet & \textbf{65.32} & \textbf{56.81} & \textbf{35.50} \\
\midrule
\multicolumn{4}{l}{\textbf{PanNuke}} \\
Method & Dice (\%) & IoU (\%) & mPQ+ (\%) \\
\midrule
PSPNet & 75.03 $\pm$ 1.11 & 71.33 $\pm$ 1.06 & --- \\
KNet & 78.46 $\pm$ 0.36 & 74.46 $\pm$ 0.27 & --- \\
Mask2Former & 66.67 $\pm$ 2.37 & 62.82 $\pm$ 2.88 & --- \\
Segformer & 69.11 $\pm$ 6.87 & 65.32 $\pm$ 7.10 & --- \\
HoVer-Net & 72.52 $\pm$ 5.45 & 68.14 $\pm$ 5.48 & 35.92 $\pm$ 0.78 \\
Zhang et al. (2024) & 73.95 $\pm$ 1.46 & 69.89 $\pm$ 1.50 & 36.06 $\pm$ 1.45 \\
HoVer-NeXt & 69.61 $\pm$ 0.57 & 65.23 $\pm$ 0.58 & 33.46 $\pm$ 0.39 \\
CellViT & 76.33 $\pm$ 0.38 & 72.04 $\pm$ 0.36 & 41.91 $\pm$ 0.30 \\
TME-ContextNet & \textbf{79.13 $\pm$ 0.80} & \textbf{74.85 $\pm$ 0.81} & \textbf{43.74 $\pm$ 2.89} \\
\bottomrule
\end{tabular}
\par\vspace{5pt}\begin{minipage}{\linewidth}\footnotesize PanNuke: mean $\pm$ sample standard deviation across three test folds. Dice and IoU follow the original pixel evaluator, including background. mPQ+ uses foreground instances with same-class matching at IoU $>0.5$. Dashes indicate that instance outputs are unavailable.\end{minipage}
\end{table}

\subsection{Class-specific behavior on PanNuke}\label{sec:classes}
Table~\ref{tab:classes} shows the five nuclear categories. TME-ContextNet has the highest Dice and IoU among the five listed methods for neoplastic, inflammatory, connective, and epithelial nuclei. Its neoplastic Dice is 78.18\%, compared with 75.49\% for CellViT and 67.67\% for HoVer-Net; its connective Dice is 56.13\%, compared with 52.03\% and 48.92\%, respectively. These class-specific results show that the overall gain is accompanied by improvements across several cell populations.

The dead-class result for Zhang et al. is an empty-prediction case. Across all three test folds, the evaluated implementation predicts no dead-class instances: both TP and FP are zero, and all 2,825 annotated dead nuclei are missed. Its dead-class instance PQ is consequently zero. The apparently high pixel Dice, IoU, and Pixel $P\times R$ values of $97.07\pm0.57$\% arise because the evaluator assigns one to an empty/empty class mask. Images without dead-cell annotations account for approximately 97\% of each fold on average and dominate these image-averaged scores. The marked row in Table~\ref{tab:classes} therefore records class absence, rather than successful recognition of dead nuclei. This case explains why the pixel measures are complemented by type-aware mPQ+ in Table~\ref{tab:overall}.

Appendix Fig.~\ref{fig:pannuke} provides additional PanNuke examples with enlarged details, and Appendix Fig.~\ref{fig:semanticdetail} illustrates a local disagreement in neoplastic and dead-cell assignments.
\begin{table}[!htbp]
\centering
\small
\caption{Per-class comparison on PanNuke. Bold indicates the highest mean within each class and metric.}
\label{tab:classes}
\setlength{\tabcolsep}{4pt}
\renewcommand{\arraystretch}{1.12}
\begin{tabular}{llccc}
\toprule
Class & Method & Dice (\%) & IoU (\%) & Pixel $P\times R$ (\%) \\
\midrule
Neoplastic & HoVer-Net & 67.67 $\pm$ 4.26 & 61.41 $\pm$ 4.33 & 60.30 $\pm$ 4.31 \\
 & Zhang et al. (2024) & 63.75 $\pm$ 5.42 & 57.39 $\pm$ 5.28 & 56.23 $\pm$ 5.31 \\
 & HoVer-NeXt & 55.04 $\pm$ 0.43 & 48.46 $\pm$ 0.84 & 47.12 $\pm$ 0.90 \\
 & CellViT & 75.49 $\pm$ 0.51 & 69.43 $\pm$ 0.12 & 68.33 $\pm$ 0.09 \\
 & TME-ContextNet & \textbf{78.18 $\pm$ 1.37} & \textbf{72.22 $\pm$ 1.06} & \textbf{71.18 $\pm$ 1.03} \\
\midrule
Inflammatory & HoVer-Net & 58.31 $\pm$ 2.14 & 53.13 $\pm$ 2.29 & 52.02 $\pm$ 2.36 \\
 & Zhang et al. (2024) & 57.75 $\pm$ 1.47 & 53.02 $\pm$ 1.40 & 52.05 $\pm$ 1.34 \\
 & HoVer-NeXt & 57.79 $\pm$ 0.60 & 53.13 $\pm$ 0.80 & 52.04 $\pm$ 0.83 \\
 & CellViT & 60.96 $\pm$ 0.22 & 55.75 $\pm$ 0.29 & 54.64 $\pm$ 0.27 \\
 & TME-ContextNet & \textbf{62.83 $\pm$ 1.75} & \textbf{57.56 $\pm$ 1.99} & \textbf{56.49 $\pm$ 2.04} \\
\midrule
Connective & HoVer-Net & 48.92 $\pm$ 1.28 & 39.42 $\pm$ 1.07 & 36.84 $\pm$ 1.07 \\
 & Zhang et al. (2024) & 46.88 $\pm$ 0.33 & 38.34 $\pm$ 0.67 & 35.95 $\pm$ 0.84 \\
 & HoVer-NeXt & 44.46 $\pm$ 1.88 & 35.39 $\pm$ 1.81 & 32.67 $\pm$ 1.86 \\
 & CellViT & 52.03 $\pm$ 0.88 & 43.04 $\pm$ 0.61 & 40.59 $\pm$ 0.51 \\
 & TME-ContextNet & \textbf{56.13 $\pm$ 0.39} & \textbf{47.09 $\pm$ 0.74} & \textbf{44.72 $\pm$ 0.89} \\
\midrule
Dead & HoVer-Net & 85.92 $\pm$ 19.86 & 85.87 $\pm$ 19.95 & 85.85 $\pm$ 19.99 \\
 & Zhang et al. (2024)\textsuperscript{*} & \textbf{97.07 $\pm$ 0.57} & \textbf{97.07 $\pm$ 0.57} & \textbf{97.07 $\pm$ 0.57} \\
 & HoVer-NeXt & 89.10 $\pm$ 4.83 & 88.94 $\pm$ 4.87 & 88.87 $\pm$ 4.89 \\
 & CellViT & 86.82 $\pm$ 2.03 & 86.66 $\pm$ 2.00 & 86.59 $\pm$ 2.00 \\
 & TME-ContextNet & 94.01 $\pm$ 5.69 & 93.85 $\pm$ 5.86 & 93.79 $\pm$ 5.93 \\
\midrule
Epithelial & HoVer-Net & 77.88 $\pm$ 5.44 & 75.78 $\pm$ 5.19 & 75.38 $\pm$ 5.15 \\
 & Zhang et al. (2024) & 81.58 $\pm$ 2.80 & 79.77 $\pm$ 2.89 & 79.38 $\pm$ 2.92 \\
 & HoVer-NeXt & 75.51 $\pm$ 2.78 & 73.47 $\pm$ 2.83 & 73.05 $\pm$ 2.86 \\
 & CellViT & 86.32 $\pm$ 0.99 & 84.23 $\pm$ 1.11 & 83.84 $\pm$ 1.13 \\
 & TME-ContextNet & \textbf{87.31 $\pm$ 2.94} & \textbf{85.13 $\pm$ 3.04} & \textbf{84.75 $\pm$ 3.07} \\
\bottomrule
\end{tabular}
\par\vspace{5pt}\begin{minipage}{\linewidth}\footnotesize Values are percentages, reported as mean $\pm$ sample standard deviation across three test folds. Pixel $P\times R$ is the product of pixel precision and recall, averaged over images. Instance-level quality is reported as mPQ+ in Table~\ref{tab:overall}. \textsuperscript{*}No dead-class nuclei were predicted in any fold. The high image-averaged pixel values reflect the empty/empty-mask score of one; dead-class instance PQ is zero.\end{minipage}
\end{table}

\subsection{Component configurations on PanMix}\label{sec:components}
The PanMix configurations are reported in Table~\ref{tab:components}. Adding CSS to the HoVer-Net configuration raises Dice from 69.80\% to 73.83\% and mPQ+ from 47.85\% to 52.48\%. The complete configuration has the highest values, reaching Dice of 80.23\%, IoU of 78.52\%, and mPQ+ of 58.67\%.

The intermediate variants clarify the roles of the architectural components. The configuration without self-attention has Dice of 77.45\% and mPQ+ of 56.19\%; removing both LGCM and SE gives 75.69\% and 54.03\%. The configuration without SE has Dice of 70.50\% and mPQ+ of 32.54\%, compared with 80.23\% and 58.67\% for the complete model. The larger difference in the joint instance score indicates that a change in feature recalibration can affect the combined detection, delineation, and type-assignment output more strongly than the pixel average. Because SE is controlled independently, its removal is reported explicitly rather than being described as a pure LGCM removal. The configuration containing CSS and attention on HoVer-Net also performs below the CSS-only configuration, indicating that attention's benefit depends on the feature-sharing arrangement in which it is used.
\begin{table}[!htbp]
\centering
\small
\caption{Component configurations on PanMix.}
\label{tab:components}
\setlength{\tabcolsep}{4pt}
\renewcommand{\arraystretch}{1.12}
\begin{tabular}{lcccccc}
\toprule
Configuration & GTD & LGCM & CSS & Dice (\%) & IoU (\%) & mPQ+ (\%) \\
\midrule
HoVer-Net & --- & --- & --- & 69.80 & 67.22 & 47.85 \\
HoVer-Net + CSS & --- & --- & $\checkmark$ & 73.83 & 71.36 & 52.48 \\
HoVer-Net + CSS + attention & --- & Partial & $\checkmark$ & 71.89 & 69.34 & 50.70 \\
w/o self-attention & $\checkmark$ & Partial & $\checkmark$ & 77.45 & 75.10 & 56.19 \\
w/o LGCM and SE & $\checkmark$ & --- & $\checkmark$ & 75.69 & 73.28 & 54.03 \\
w/o SE & $\checkmark$ & $\checkmark$ & $\checkmark$ & 70.50 & 68.18 & 32.54 \\
Full model & $\checkmark$ & $\checkmark$ & $\checkmark$ & \textbf{80.23} & \textbf{78.52} & \textbf{58.67} \\
\bottomrule
\end{tabular}
\par\vspace{5pt}\begin{minipage}{\linewidth}\footnotesize Values are percentages. Partial denotes the attention-only or dilation-only LGCM configuration identified by the row label. SE is an independent switch; its removal is stated explicitly. mPQ+ uses foreground instances and IoU $>0.5$.\end{minipage}
\end{table}

\subsection{Spatial target controls on a HoVer-Net-based model}\label{sec:targets}
Table~\ref{tab:targets} compares CSS target constructions in the dedicated HoVer-Net-based control model. The experiment examines the auxiliary target while holding the control architecture fixed. EDT gives the highest mean Dice and IoU among the tested constructions, reaching 53.60\% and 45.50\%, respectively. Compared with the no-CSS condition (51.91\% Dice and 43.60\% IoU), these values represent improvements of 1.69 and 1.90 percentage points. They support the use of the class-conditioned distance target within this control experiment.

\begin{figure}[tbp]
\centering\includegraphics[width=\linewidth]{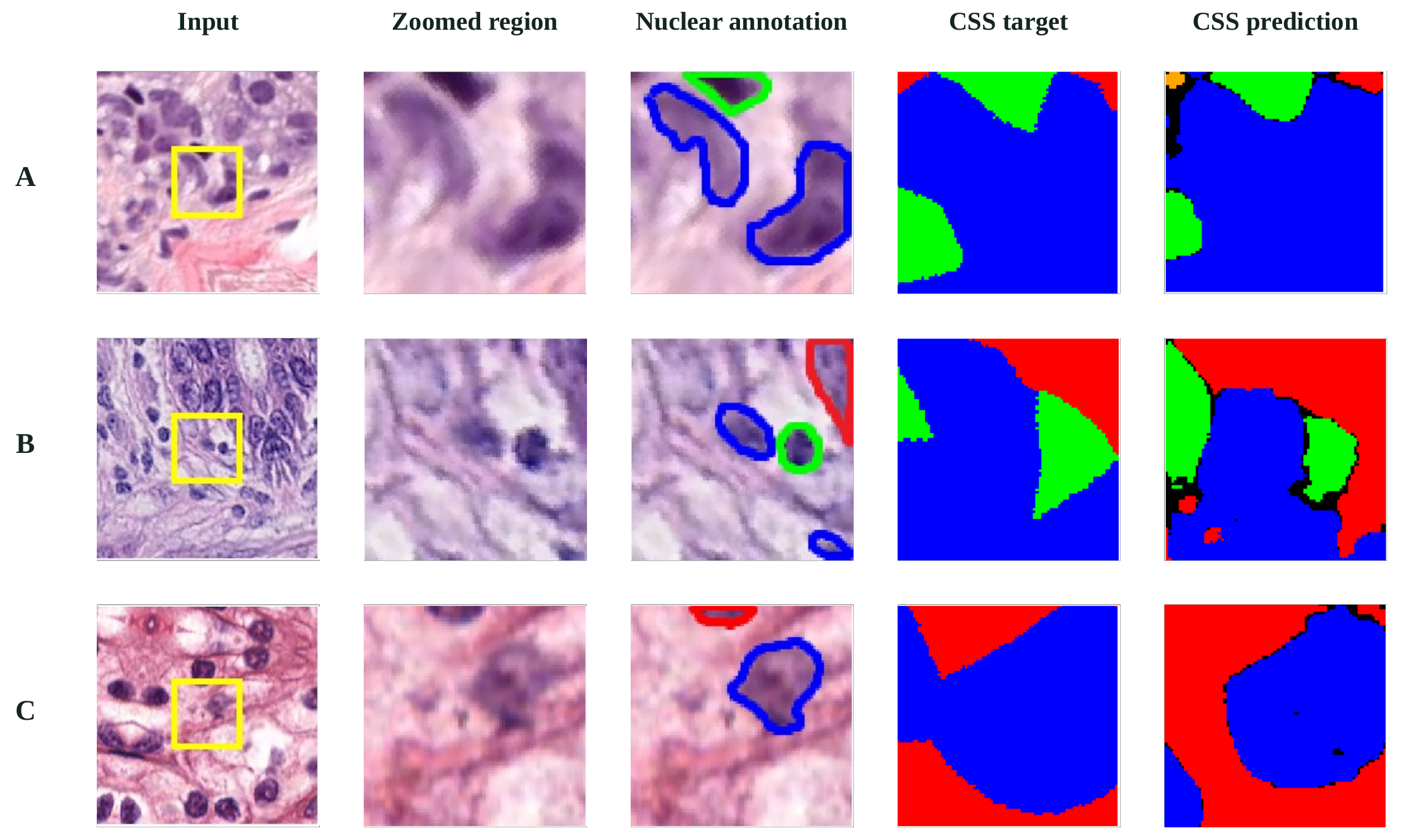}
\caption{Examples of spatial supervision targets and predictions. From left to right, each row shows the input image, enlarged region of interest, nuclear annotations, annotation-derived CSS target, and predicted CSS classes. The spatial target extends class assignments into non-nuclear regions while retaining the annotated nuclear labels}
\label{fig:cssprediction}
\end{figure}

Uniform Gaussian, class-weighted Gaussian, constant class weights, and permuted class weights all yield lower Dice and IoU than the no-CSS condition. Permuted class weights yield 51.32\% Dice and 41.47\% IoU. Class-agnostic distance also performs below the no-CSS condition, with 49.39\% Dice and 41.08\% IoU; EDT exceeds this construction by 4.21 and 4.42 percentage points. The results favor extending class-specific nuclear neighborhoods through EDT over the tested Gaussian and type-agnostic alternatives. They also show that the benefit of auxiliary spatial supervision depends on its target definition. These controls distinguish the use of class information from the mere presence of an auxiliary distance task. Removing class identity changes what the model is asked to encode, while class weights and Gaussian accumulation alter the spatial assignments. Their observed scores support the particular EDT construction in this control setting. The alternatives trained with one seed provide descriptive comparisons rather than a statistical ranking. Figure~\ref{fig:cssprediction} illustrates the spatial regions learned by the auxiliary task.

\begin{table}[!htbp]
\centering
\small
\caption{CSS target constructions in a dedicated HoVer-Net-based auxiliary-supervision model on CoNSeP. EDT is the distance-transform construction adopted for CSS.}
\label{tab:targets}
\setlength{\tabcolsep}{4pt}
\renewcommand{\arraystretch}{1.12}
\begin{tabular}{lccc}
\toprule
Target & Seeds & Dice (\%) & IoU (\%) \\
\midrule
No CSS & 3 & 51.91 $\pm$ 3.30 & 43.60 $\pm$ 3.15 \\
EDT & 3 & 53.60 $\pm$ 1.30 & 45.50 $\pm$ 0.99 \\
Uniform Gaussian & 1 & 50.45 & 42.49 \\
Class-weighted Gaussian & 1 & 48.86 & 40.81 \\
Constant class weights & 1 & 51.26 & 42.73 \\
Permuted class weights & 1 & 51.32 & 41.47 \\
Class-agnostic distance & 1 & 49.39 & 41.08 \\
\bottomrule
\end{tabular}
\par\vspace{5pt}\begin{minipage}{\linewidth}\footnotesize No CSS and EDT: mean $\pm$ sample standard deviation across three seeds. Other targets: one seed. The control architecture is held fixed across targets. Values are percentages and follow the original pixel evaluator.\end{minipage}
\end{table}

\subsection{Decoder organization and LGCM}\label{sec:structures}
Table~\ref{tab:structures} isolates the effect of decoder sharing depth on PanMix and reports the dedicated CoNSeP decoder and LGCM controls. The controlled PanMix comparison disables CSS, SE, and self-attention throughout and varies only the shared decoding stages. Stages I, II, and III use 40.60, 36.80, and 35.31 million parameters, respectively. Stage II achieves the highest Dice and IoU, at 76.00\% and 73.61\%. Extending sharing from u3 to u3 and u2 improves overlap, whereas additionally sharing u1 lowers overlap despite a further reduction in parameters. These results support sharing intermediate decoder features while retaining task-specific final refinement.

The controlled CoNSeP decoder blocks show that the preferred arrangement depends on auxiliary supervision. Without CSS, the independent decoder has the highest Dice, at 56.55\%, followed by the plain shared decoder and GTD. With CSS, GTD reaches Dice of 54.98\%, compared with 51.72\% for the independent decoder, while using 36.40 million rather than 65.18 million parameters, a reduction of 44.2\%. This combination of higher overlap and reduced decoder duplication supports pairing GTD with CSS.

LGCM has a positive effect in both dedicated blocks. With SE disabled, adding LGCM to the HoVer-Net control raises Dice from 47.47\% to 56.61\% and IoU from 39.65\% to 48.19\%. In the CoNSeP GTD control with SE enabled, the effect is smaller: Dice is 54.98\% with LGCM and 54.33\% without it. The two blocks indicate that LGCM can supply useful context independently, while its marginal contribution changes when the remaining components are present.
\begin{table}[!htbp]
\centering
\small
\caption{Controlled comparisons of decoder sharing depth, decoder organization, and LGCM.}
\label{tab:structures}
\setlength{\tabcolsep}{4pt}
\renewcommand{\arraystretch}{1.12}
\begin{tabular}{lccc}
\toprule
Configuration & Params (M) & Dice (\%) & IoU (\%) \\
\midrule
\multicolumn{4}{l}{\textit{PanMix: controlled sharing depth}} \\
HoVer-Net & 54.74 & 69.80 & 67.22 \\
Integration Stage I & 40.60 & 74.33 & 71.86 \\
Integration Stage II & 36.80 & 76.00 & 73.61 \\
Integration Stage III & 35.31 & 66.28 & 64.03 \\
\midrule
\multicolumn{4}{l}{\textit{CoNSeP: decoder / no CSS}} \\
Independent decoder & 65.18 & 56.55 & 47.87 \\
Plain shared decoder & 35.99 & 54.66 & 46.88 \\
GTD & 36.40 & 53.75 & 45.33 \\
\midrule
\multicolumn{4}{l}{\textit{CoNSeP: decoder / with CSS}} \\
GTD + CSS control & 36.40 & 54.98 & 46.62 \\
Independent decoder & 65.18 & 51.72 & 43.43 \\
\midrule
\multicolumn{4}{l}{\textit{CoNSeP: LGCM only / SE off}} \\
HoVer-Net  & 54.74 & 47.47 & 39.65 \\
HoVer-Net + LGCM & 54.79 & 56.61 & 48.19 \\
\midrule
\multicolumn{4}{l}{\textit{CoNSeP: LGCM in GTD / SE on}} \\
GTD control & 36.40 & 54.98 & 46.62 \\
GTD control w/o LGCM & 36.38 & 54.33 & 45.96 \\
\bottomrule
\end{tabular}
\par\vspace{5pt}\begin{minipage}{\linewidth}\footnotesize Params: million parameters. Accuracy values are percentages. The no-CSS and with-CSS blocks evaluate decoder organization under their stated supervision. LGCM-only comparisons disable SE; the GTD control comparison enables SE. The CoNSeP rows are dedicated control runs with seed 10. PanMix Stages I--III vary only the shared decoding depth. CSS, SE, and self-attention are disabled in all three stages; all remaining training and evaluation settings are held fixed.\end{minipage}
\end{table}

\subsection{Auxiliary-weight sensitivity on PanMix}\label{sec:sensitivity}
\begin{figure}[tbp]
\centering\includegraphics[width=\linewidth]{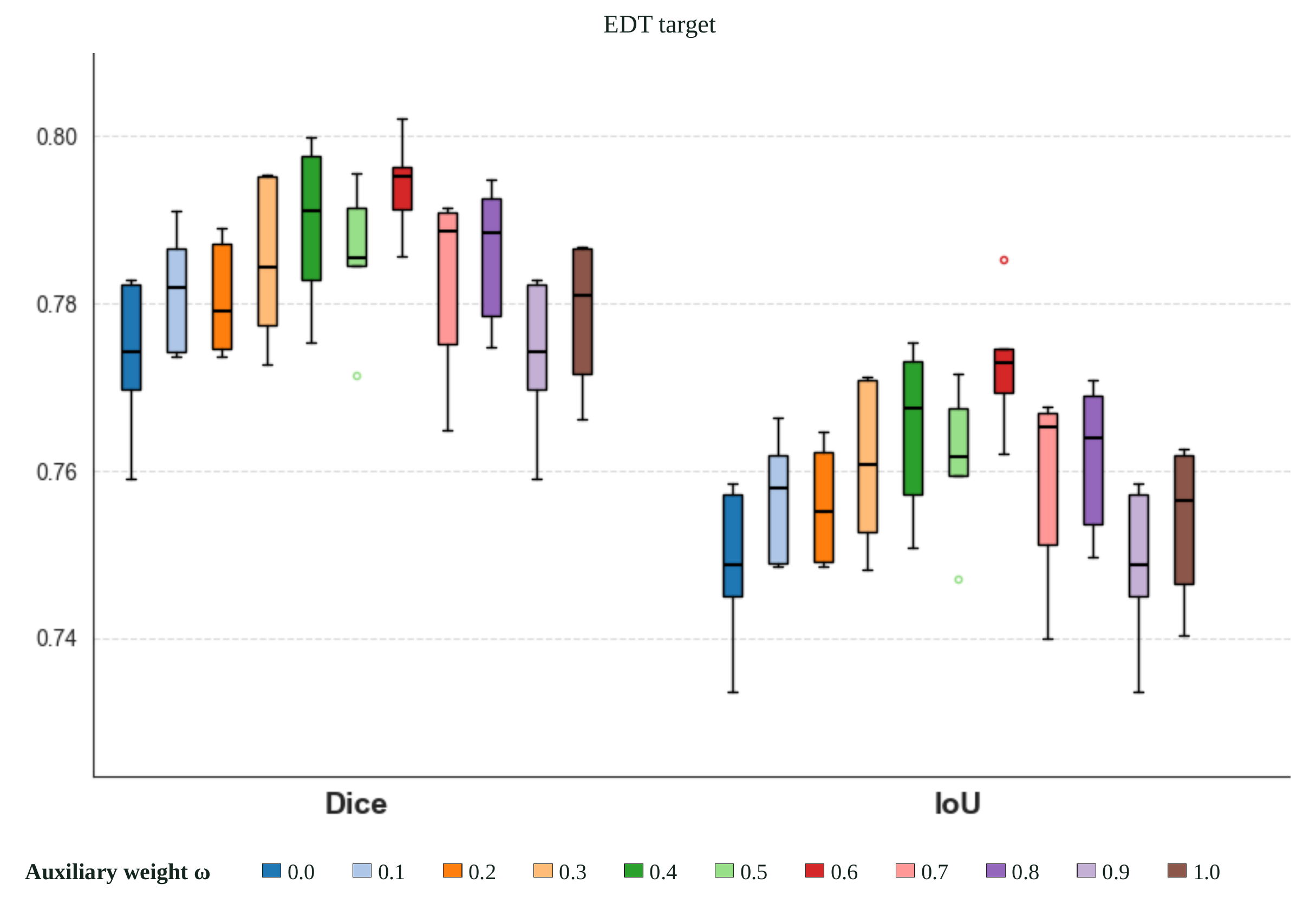}
\caption{Effect of auxiliary-loss weight $\omega$ on PanMix with the EDT-based CSS target. Colors denote weights from 0.0 to 1.0 in increments of 0.1. Each weight uses one training run with the remaining settings held fixed. The boxplots summarize per-image test scores, rather than variation across random seeds. Scores are shown as fractions.}
\label{fig:omega}
\end{figure}

Figure~\ref{fig:omega} examines how the strength of the EDT auxiliary objective changes overlap on PanMix. The recorded boxplots show higher central Dice and IoU values for moderate weights, with $\omega=0.6$ giving the highest visible medians among the examined settings. Increasing the weight beyond this range does not yield a monotonic improvement. The pattern is consistent with auxiliary supervision being useful when it contributes to representation learning without dominating the primary tasks.

This observation complements the target controls. A target construction determines the auxiliary gradient direction, while $\omega$ changes the magnitude of its contribution. Equation~\eqref{eq:gradient} separates these two effects through the inner-product term. The sensitivity experiment shows how changing the auxiliary-loss weight affects the observed overlap, complementing the optimization role of $\omega$ in Eq.~\eqref{eq:gradient}.

\subsection{Computational cost}\label{sec:efficiency}
Table~\ref{tab:efficiency} reports model size, arithmetic cost, latency, and peak GPU memory. In the RTX 4090 main-model session, TME-ContextNet uses 36.40 million parameters and 267.77 GFLOPs, compared with 54.74 million and 398.32 GFLOPs for HoVer-Net. These values reduce parameters by 33.5\% and arithmetic cost by 32.8\%. Its mean network latency is 19.37 ms and its mean image latency is 357.24 ms, compared with 24.78 ms and 444.17 ms for HoVer-Net. The CSS and no-CSS rows have the same listed parameter count and arithmetic cost, consistent with auxiliary supervision being confined to training. Both retain the same inference prediction path.

The RTX 4090 decoder blocks provide a separate comparison of sharing arrangements. In the block that includes all three decoder structures, GTD and the plain shared decoder use fewer parameters, fewer operations, and less peak memory than the independent decoder. The plain shared decoder has the lower forward latency, while GTD has the lower image latency in that block. Repeated timings in the two 4090 sessions vary, so the measurements are presented in their original blocks. Network and image latency are complementary: post-processing and image assembly account for a substantial part of the latter.
\begin{table}[!htbp]
\centering
\footnotesize
\caption{Computational cost of TME-ContextNet and decoder variants on RTX 4090. Measurement sessions are presented separately.}
\label{tab:efficiency}
\setlength{\tabcolsep}{4pt}
\renewcommand{\arraystretch}{1.12}
\begin{tabular}{lrrrrrr}
\toprule
Method & Params & GFLOPs & Network & Image & P95 & Memory \\
 & (M) & & (ms) & (ms) & (ms) & (GiB) \\
\midrule
\multicolumn{7}{l}{\textit{RTX 4090: main-model session}} \\
HoVer-Net & 54.74 & 398.32 & 24.78 & 444.17 & 463.02 & 0.50 \\
TME-ContextNet & 36.40 & 267.77 & 19.37 & 357.24 & 410.72 & 0.43 \\
TME-ContextNet w/o CSS & 36.40 & 267.77 & 19.21 & 351.31 & 381.46 & 0.43 \\
\midrule
\multicolumn{7}{l}{\textit{RTX 4090: session 1}} \\
GTD / no CSS & 36.40 & 267.77 & 21.33 & 388.49 & 397.46 & 0.43 \\
Plain shared / no CSS & 35.99 & 262.53 & 20.84 & 385.69 & 412.30 & 0.43 \\
\midrule
\multicolumn{7}{l}{\textit{RTX 4090: session 2}} \\
GTD / no CSS & 36.40 & 267.77 & 19.02 & 339.07 & 341.82 & 0.43 \\
Plain shared / no CSS & 35.99 & 262.53 & 18.75 & 345.39 & 385.64 & 0.43 \\
Independent / no CSS & 65.18 & 410.79 & 33.23 & 587.62 & 607.95 & 0.54 \\
\bottomrule
\end{tabular}
\par\vspace{5pt}\begin{minipage}{\linewidth}\footnotesize FP32, batch size one. Network: mean forward latency. Image: mean latency including post-processing and excluding disk input/output for a $256\times256$ image. P95: 95th-percentile image latency. GFLOPs: one native network input patch. Memory: peak GPU memory. The main-model block was remeasured on October 28. Decoder sessions 1 and 2 retain the October 21 and October 22 measurements, respectively. Models within each block were timed sequentially in the same GPU session.\end{minipage}
\end{table}

\section{Discussion}\label{sec:discussion}
\subsection{Spatial supervision for quantitative cellular analysis}
TME-ContextNet combines dense spatial supervision with grouped decoding to address the different contextual requirements of nuclear delineation and type recognition. The complete framework has the highest overall pixel and instance scores among the evaluated methods across PanMix, CoNSeP, and PanNuke. Improvements across several PanNuke cell populations complement this overall comparison. The computational results show that the joint outputs can be obtained with fewer parameters and less arithmetic than the HoVer-Net comparator, making the design relevant to quantitative cellular analysis in imaging informatics.

CSS makes a specific part of the annotation information available to representation learning. Nuclear type labels identify cell populations at annotated nuclear locations. Their distance-based extension additionally organizes the surrounding space according to proximity to those populations. This connects the biological motivation of tissue organization to a reproducible geometric target. The nearest-region interpretation and perturbation analysis explain what that target represents and how its assignments respond to small changes in nuclear support. This spatial partition provides dense contextual supervision for the joint nuclear outputs.

The existing controls support that interpretation at two levels. The PanMix component comparison shows higher joint metrics when CSS is added to the HoVer-Net baseline, and the dedicated HoVer-Net-based CoNSeP target controls show higher pixel overlap for EDT than for no CSS. Across the tested target constructions, EDT has the highest Dice and IoU, whereas the Gaussian and class-agnostic alternatives perform below the no-CSS condition. These results favor the class-conditioned EDT construction and show that the spatial target definition influences the benefit of auxiliary learning. Equation~\eqref{eq:gradient} explains how different targets can alter shared-parameter updates through their alignment with the primary objective. It explains the contribution of the auxiliary objective to shared-feature optimization.

For imaging informatics, the useful output is a population of individually delineated and typed nuclei from which cellular summaries can be derived. The overall mPQ+ results address this joint output more directly than class-map overlap alone. The inclusion of general segmentation methods establishes a demanding pixel-level reference, while the specialist comparison tests whether that representation also supports typed instances. These results demonstrate the potential of TME-ContextNet as an efficient component for quantitative cellular analysis in computational pathology.

\subsection{Interaction between supervision and task sharing}
GTD addresses the accompanying feature-sharing problem. Shared decoding is established in nuclei analysis, including the enhanced framework of Zhang et al. \cite{zhang2024keep}. The present design groups the late representations by geometry and semantics and pairs CSS with nuclear type prediction. Its advantage is clearest in the with-CSS comparison, where it combines higher Dice with fewer parameters than independent decoding. Independent decoding has higher Dice in the no-CSS block. These observations support the coupling of grouped decoding and spatial supervision as the relevant design choice, with a clear accuracy--cost trade-off. The controlled PanMix results also favor intermediate sharing: Stage II shares u3 and u2 but retains task-specific u1 refinement and outperforms both shallower sharing and sharing through u1. With the other components held fixed, this comparison identifies the depth of sharing as a distinct decoder design choice.

LGCM provides complementary context through encoder dilation and decoder self-attention. The dedicated SE-off comparison shows that LGCM can improve the simpler architecture on its own, while the GTD control with SE enabled shows a smaller positive increment. The difference in marginal benefit is consistent with contextual information being supplied by several parts of the network. The ablations support complementary contributions from contextual features and grouped spatial supervision.

The PanMix weight analysis provides a related optimization perspective. Its non-monotonic trend indicates that increasing the amount of auxiliary supervision is not sufficient by itself. The finite-step argument explains why auxiliary direction and weight both enter the update, while the experiments show the behavior of specific targets and settings. The gradient analysis explains the optimization role of the auxiliary objective, while the weight experiment characterizes its empirical behavior.

\subsection{Computational implications and further development}
\begin{figure}[tbp]
\centering\includegraphics[width=\linewidth]{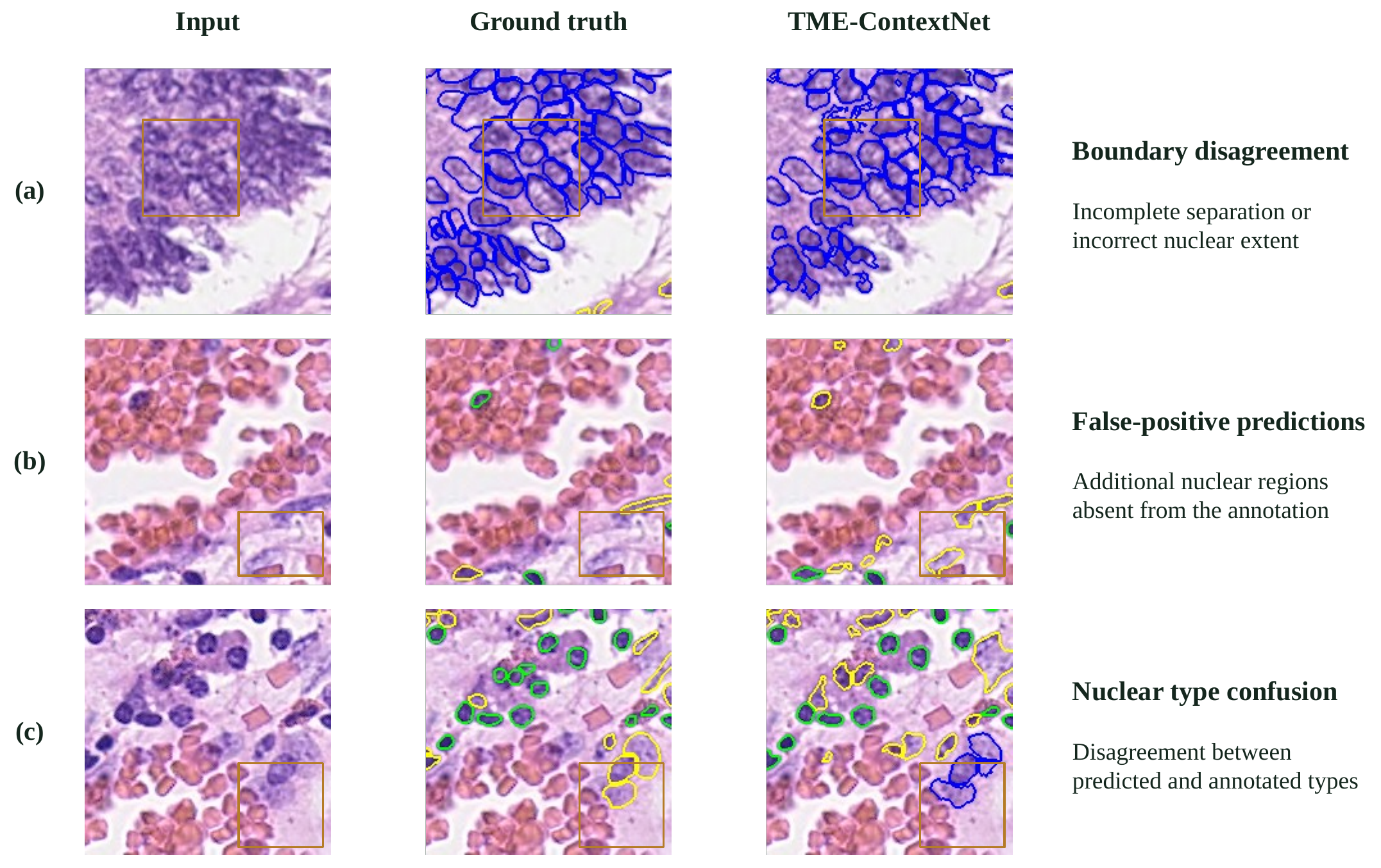}
\caption{Examples of residual prediction errors on CoNSeP. Each row shows the input, ground truth, and TME-ContextNet output. Outlined regions highlight (a) incomplete separation or disagreement in nuclear extent, (b) additional predicted nuclear regions absent from the annotation, and (c) disagreement between predicted and annotated nuclear types. Nuclear contour colors follow the CoNSeP legend in Fig.~\ref{fig:comparison}}
\label{fig:failures}
\end{figure}

Grouped decoding reduces redundant computation in repeated cellular analysis. The image-level timings identify tiling and instance assembly as substantial additional processing costs, making efficient assembly a useful companion to network acceleration. Whole-slide throughput also depends on tile scheduling and batching. The challenging examples in Fig.~\ref{fig:failures} identify three concrete development targets: nuclear extent and separation, false-positive suppression, and nuclear type discrimination. Adaptive spatial supervision and efficient image-level assembly offer complementary directions for improving these outputs and their processing cost.

\section{Conclusions}\label{sec:conclusion}
TME-ContextNet combines a class-conditioned distance target, hierarchical local--global features, and grouped multi-task decoding for joint nuclei instance segmentation and classification. The mathematical formulation connects the implemented auxiliary field to a nearest-class spatial partition and explains how its training signal reaches shared features. Across the three evaluated datasets, the complete framework achieves the highest overall Dice, IoU, and mPQ+ values, while the computational measurements show reduced cost relative to the evaluated HoVer-Net implementation. Target and decoder controls identify the conditions supporting these gains, providing a practical basis for further development of context-aware cellular analysis in computational pathology.

\FloatBarrier
\appendix
\section{Derivations for spatial supervision and optimization}\label{app:derivation}
\subsection{Nearest-class assignment and numerical clipping}
For any two present classes $a$ and $b$ whose scores are not clipped,
\begin{equation}
 \log S_a(p)-\log S_b(p)=\frac{d_b(p)^2-d_a(p)^2}{2\sigma^2}.
\end{equation}
Thus $S_a(p)>S_b(p)$ exactly when $d_a(p)<d_b(p)$, proving Eq.~\eqref{eq:nearest} for a unique, unclipped winner. The common scale cancels in the ordering. If every class distance is at least $\sigma\sqrt{2\log(1/\varepsilon)}$, all scores equal the floor and the implementation's ordered tie rule determines the assigned class. This threshold identifies the far-field region governed by the ordered tie rule.

For comparison, an unclipped class-weighted score $S_c=w_c\exp[-d_c^2/(2\sigma_c^2)]$ has assignment rule
\begin{equation}
 \argmax_c S_c=\argmin_c\left\{\frac{d_c^2}{2\sigma_c^2}-\log w_c\right\}.
\end{equation}
Weights and class-specific scales move spatial decision boundaries. A Gaussian sum over multiple nuclear pixels is a different construction: it can also change with the number and distribution of contributing pixels. These distinctions explain why the EDT, Gaussian, and weight-control conditions are treated as different target designs.

\subsection{Perturbation of class supports}
Assume nonempty supports $F_c$ and $F'_c$ satisfy $d_H(F_c,F'_c)\leq\delta$, where $d_H$ is their Hausdorff distance. For a nearest point $q\in F_c$, there is a point $q'\in F'_c$ with $\|q-q'\|\leq\delta$. The triangle inequality gives
\begin{equation}
 d'_c(p)\leq\|p-q'\|\leq\|p-q\|+\delta=d_c(p)+\delta.
\end{equation}
Interchanging the two supports proves $|d'_c(p)-d_c(p)|\leq\delta$. If the initial winning class is $a$ and every other class $b$ satisfies $d_b(p)-d_a(p)>2\delta$, then
\begin{equation}
 d'_b(p)-d'_a(p)\geq d_b(p)-d_a(p)-2\delta>0.
\end{equation}
Hence the nearest class is unchanged under the perturbation. This statement assumes the same set of present classes and excludes target pixels relabeled as nuclear foreground and score-floor ties. Under these conditions, the distance margin determines the robustness of the spatial assignment.

\subsection{A finite-step form of the auxiliary-gradient argument}
If the primary loss is $L$-smooth in the shared parameters, the descent lemma bounds the update in Eq.~\eqref{eq:gradient} by
\begin{equation}
 \mathcal L_{\rm main}(\theta')-\mathcal L_{\rm main}(\theta)
 \leq-\eta\left(\|g\|^2+\omega\langle g,a\rangle\right)
 +\frac{L\eta^2}{2}\|g+\omega a\|^2.
\end{equation}
If $B=\|g\|^2+\omega\langle g,a\rangle>0$ and $g+\omega a\ne0$, the bound is negative for $0<\eta<2B/(L\|g+\omega a\|^2)$. This bound relates the allowable step size to the auxiliary weight and target-induced gradient direction, explaining their joint role in primary-loss descent.

\section{Definitions of the CoNSeP target controls}\label{app:controls}
The Gaussian controls use an unnormalized sum over nuclear pixels,
\begin{equation}
 G_c(p)=w_c\sum_{q\in F_c}\exp\left[-\frac{\|p-q\|_2^2}{2\sigma_c^2}\right].
\end{equation}
The largest score assigns each background pixel; nuclear labels remain unchanged. In class-ID order $(1,2,3,4)$, Uniform Gaussian uses $w=(1,1,1,1)$ and $\sigma=(5,5,5,5)$ pixels. Class-weighted Gaussian uses $w=(1,1,2,2)$ and $\sigma=(5,5,8,8)$. Constant class weights uses $w=(1,1,1,1)$, and Permuted class weights uses $w=(2,2,1,1)$, both retaining $\sigma=(5,5,8,8)$. Far-field assignments use the log domain where needed to avoid numerical underflow.

Class-agnostic distance uses $d(p)=\min_{q\in F}\|p-q\|_2$ for the union of nuclei $F=\{p:I(p)>0\}$. Its five labels encode $d=0$, $0<d\leq4$, $4<d\leq8$, $8<d\leq16$, and $d>16$ pixels; nucleus-free images receive the farthest band. Thus, the target supervises geometric distance bands. No CSS disables the auxiliary loss; EDT follows Eqs.~\eqref{eq:distance}--\eqref{eq:target}.

\clearpage
\section{Additional architecture and qualitative illustrations}\label{app:figures}
\begin{figure}[htbp]
\centering\includegraphics[width=\linewidth]{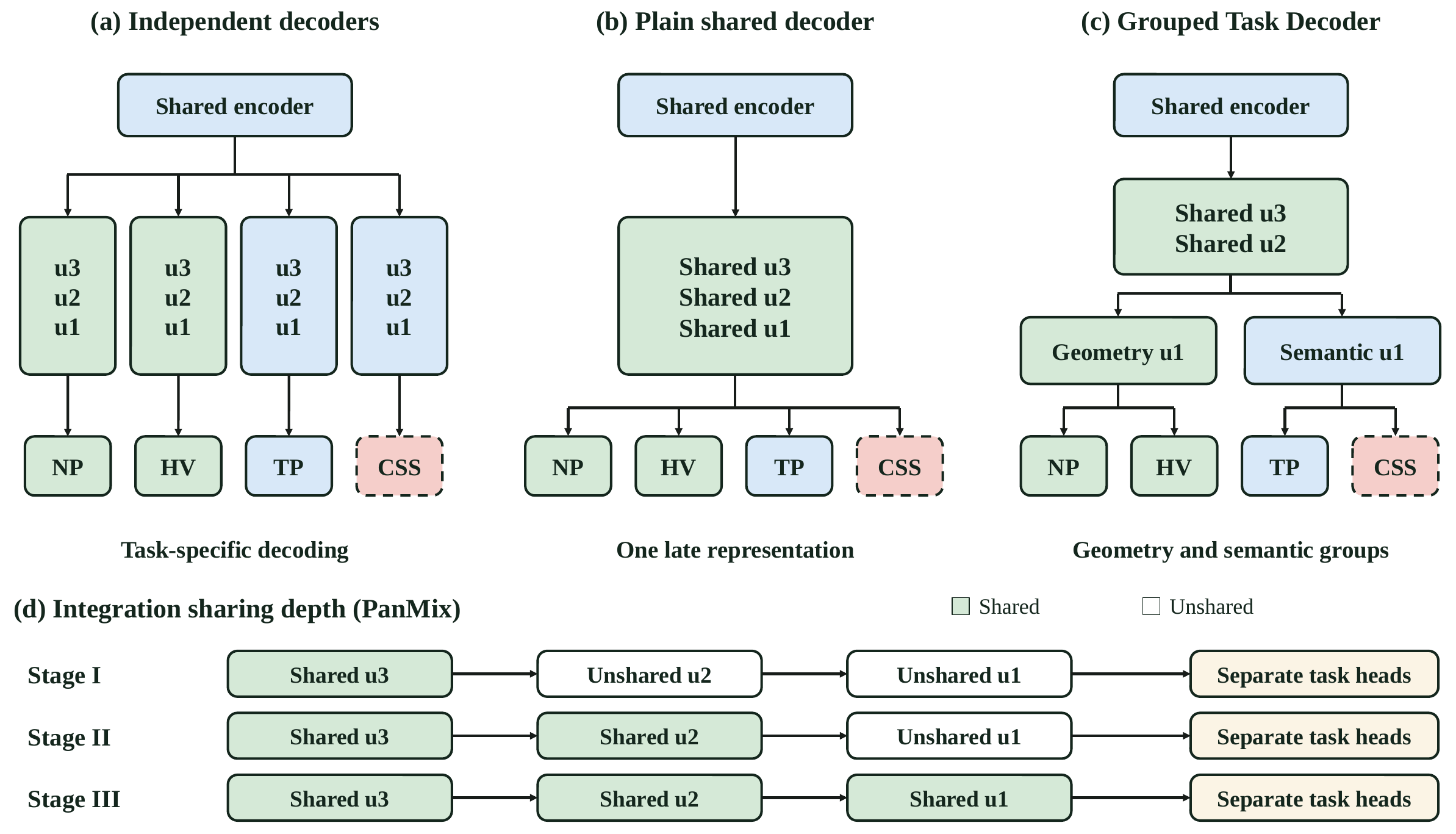}
\caption{Decoder organization and shared decoding depth. (a) Independent decoders process NP, HV, TP, and CSS separately. (b) A plain shared decoder uses one representation through stages u3, u2, and u1 before separate task heads. (c) GTD shares u3 and u2, then separates u1 refinement into geometry-related (NP/HV) and semantic (TP/CSS) groups. (d) In the controlled PanMix sharing-depth comparison in Table~\ref{tab:structures}, Stage I shares u3, Stage II shares u3 and u2, and Stage III shares u3, u2, and u1. CSS, SE, and self-attention are disabled in all three stages, and all other settings are fixed. The NP, HV, and TP task heads remain separate}
\label{fig:merging}
\end{figure}

\begin{figure}[htbp]
\centering\includegraphics[width=\linewidth]{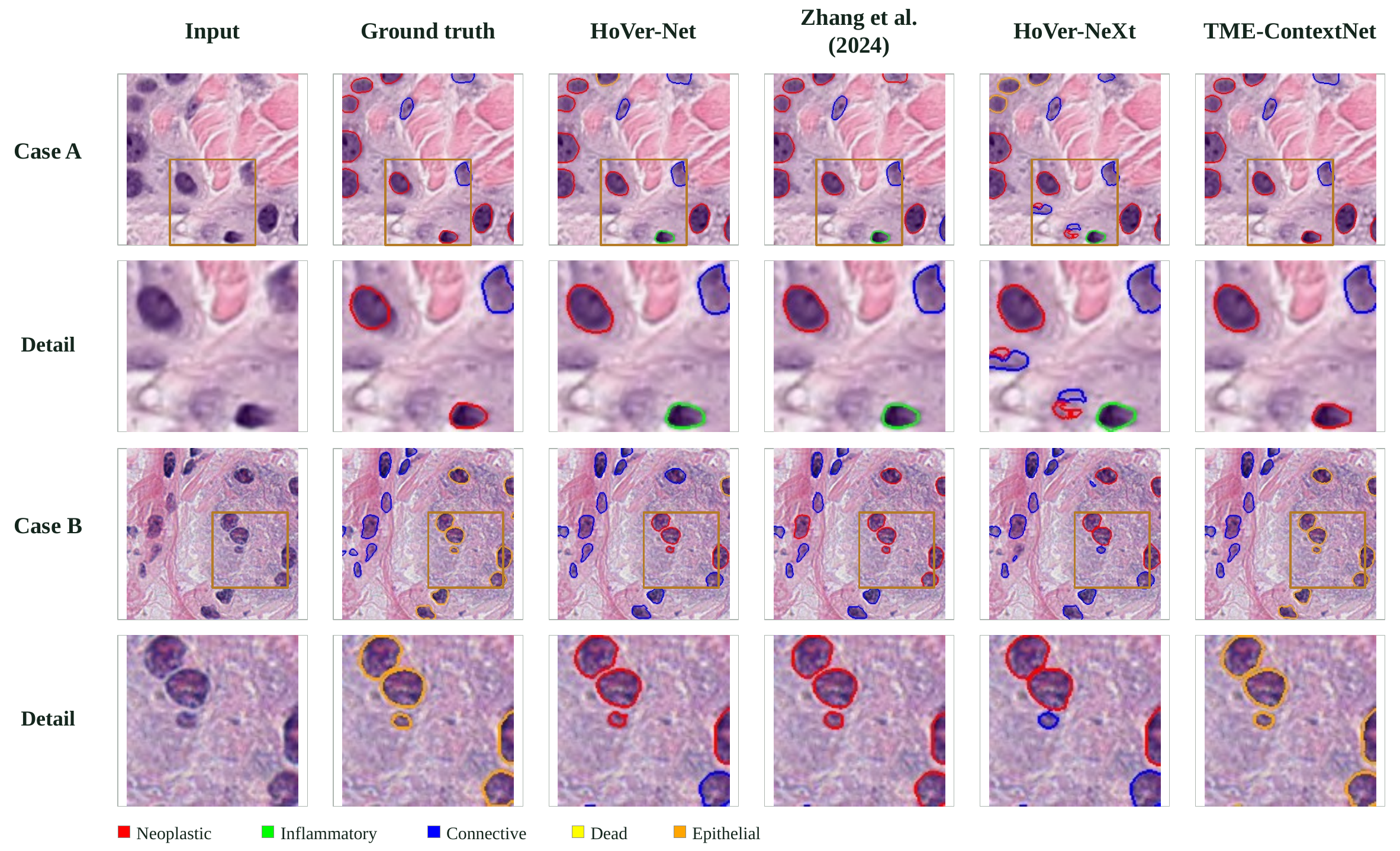}
\caption{PanNuke examples with enlarged details. Columns show the H\&E input, ground truth, HoVer-Net, Zhang et al. (2024), HoVer-NeXt, and TME-ContextNet. Each overview is followed by an enlarged view of the outlined region. All methods use the same image coordinates and class colors. The examples illustrate nuclear delineation and type assignments for the five PanNuke categories}
\label{fig:pannuke}
\end{figure}
\clearpage

\section{Additional comparisons with general segmentation methods}\label{app:semantic}
The following illustrations complement the specialist comparison in Fig.~\ref{fig:comparison}. General segmentation methods provide nuclear class regions; their displayed contours should be interpreted as boundaries of those class regions. The same H\&E field and class colors are retained across methods in each row.

\begin{figure}[htbp]
\centering\includegraphics[width=\linewidth]{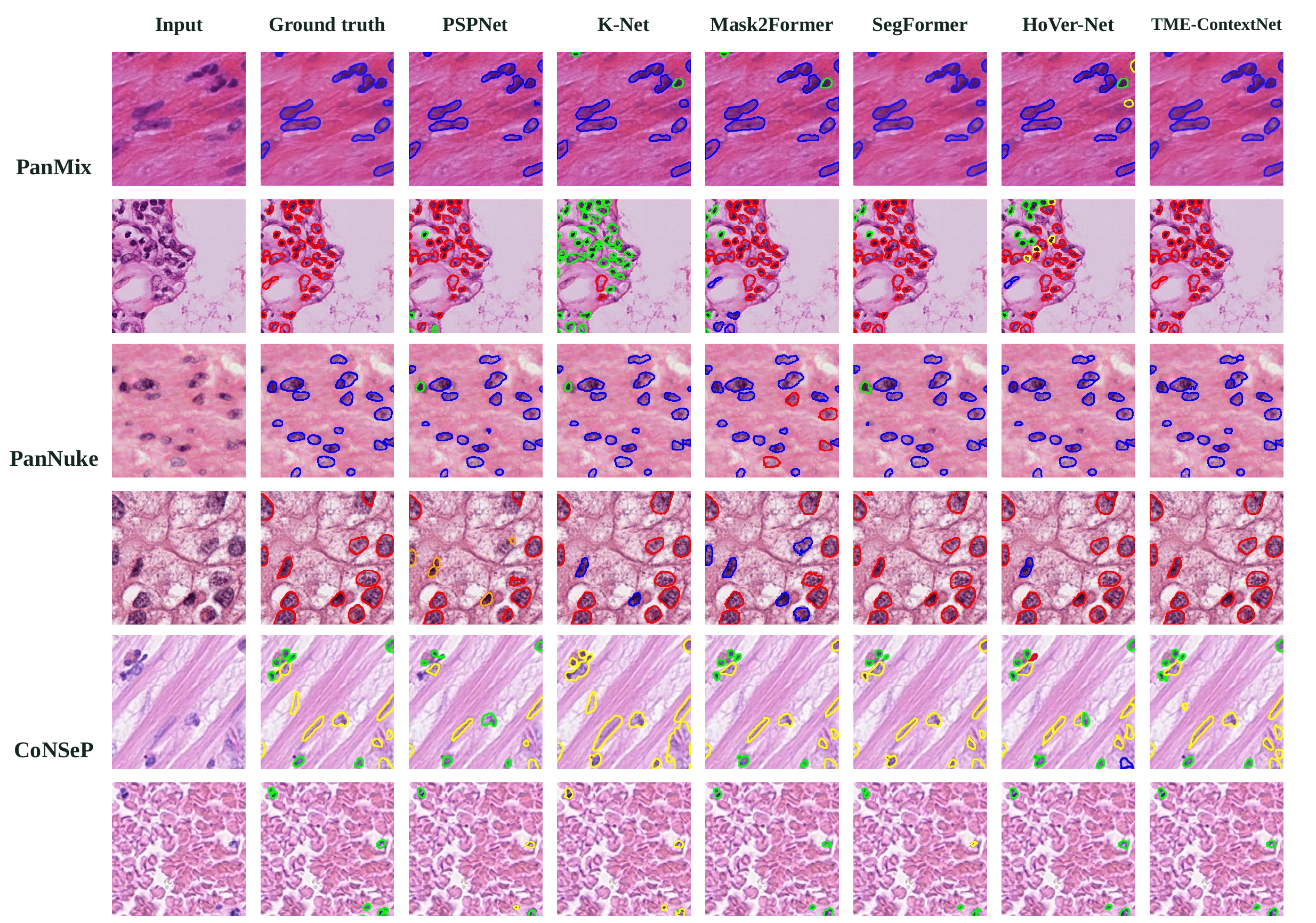}
\caption{Additional qualitative class-map comparisons on PanMix, PanNuke, and CoNSeP. Two examples per dataset are shown for the H\&E input, ground truth, PSPNet, K-Net, Mask2Former, SegFormer, HoVer-Net, and TME-ContextNet. These examples compare local class assignments and region boundaries. All methods use the same image coordinates within each example}
\label{fig:semanticcomparison}
\end{figure}

\begin{figure}[htbp]
\centering\includegraphics[width=\linewidth]{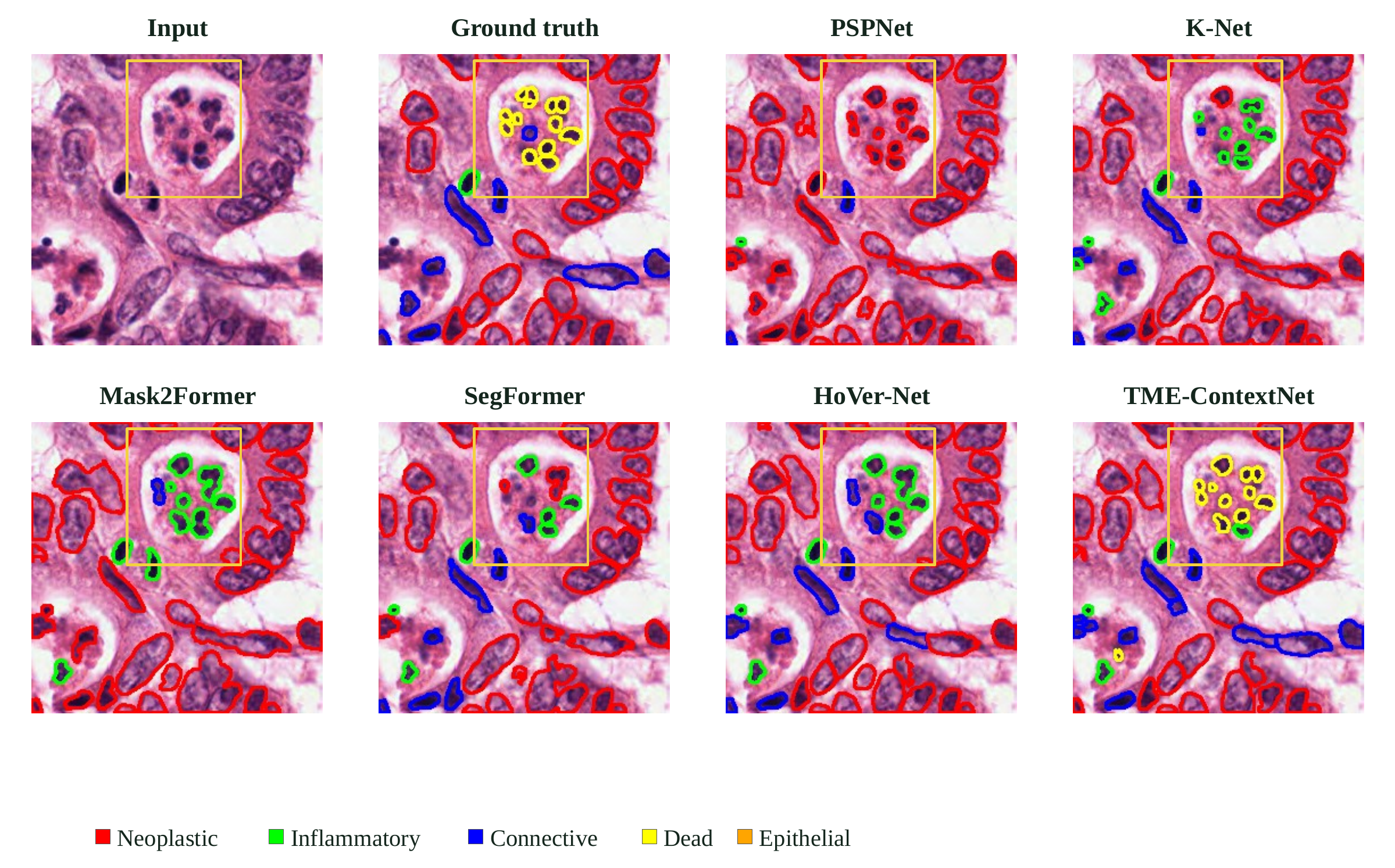}
\caption{PanNuke example of local nuclear type assignments. The outlined region identifies the same group of nuclei across the input, ground truth, and six method outputs. The example includes disagreement between neoplastic and dead-cell labels among methods. It illustrates local type ambiguity and complements the class-wise quantitative results}
\label{fig:semanticdetail}
\end{figure}
\clearpage

\FloatBarrier
\section*{Declarations}
\paragraph{Funding.} This work was supported by the Ministry of Science and Technology of China under Grant SQ2020YFA0713800 and the National Natural Science Foundation of China under Grants 12001273 and 12201109.
\paragraph{Competing interests.} The authors declare that they have no competing interests.
\paragraph{Ethics and consent.} The pathological images were anonymized before analysis. No identifiable individual participant information is presented.
\paragraph{Data and code availability.} CoNSeP and PanNuke are available from their respective public sources \cite{graham2019hover,gamper2019pannuke,gamper2020pannuke}. Source-code access and PanMix research data requests are managed through the project page: \url{https://github.com/rukang5413-jpg/TME-ContextNet}.

\end{document}